\documentclass[11pt]{article}

\usepackage[preprint]{acl}

\usepackage{times}
\usepackage{latexsym}

\usepackage[T1]{fontenc}
\usepackage[utf8]{inputenc}

\usepackage{microtype}

\usepackage{inconsolata}

\usepackage{graphicx}

\usepackage{amsmath}
\usepackage{dsfont}
\usepackage{expex}
\usepackage{array}
\usepackage[dvipsnames]{xcolor}
\usepackage{dsfont}
\usepackage[acronym, toc]{glossaries}
\RequirePackage{silence}
\usepackage{fancybox,framed}
\usepackage{array}
\usepackage{tabularray}
\usepackage{multirow}
\usepackage{wrapfig}
\usepackage{framed}
\usepackage{amssymb}
\usepackage{amsmath}
\usepackage{amsthm}
\usepackage{stmaryrd}
\usepackage{framed}
\usepackage{todonotes}
\reversemarginpar
\presetkeys{todonotes}{size=\tiny, prepend, caption={TODO}}{}
\usepackage{listings}
\usepackage{longtable}
\usepackage{multicol}
\usepackage{verbatim}  
\usepackage{verbatimbox}
\usepackage{enumitem}
\usepackage{adjustbox}
\usepackage{hyperref}
\usepackage{pifont}
\usepackage{xcolor}
\usepackage{tabularx}
\usepackage{booktabs}
\usepackage{subcaption}
\usepackage{fontawesome5}

\usepackage{expex}
\title{Separating Syntax from Language: A Mechanistic Account of Translation in Multilingual LLMs}
\newcommand{\affilsup}[1]{\rlap{\textsuperscript{\normalfont#1}}}

\author{
    Mikhail Sonkin\affilsup{1,4\textbf{*}} \qquad 
    Tanja Baeumel\affilsup{1,2,3\textbf{*}} \\
    \textbf{Daniil Gurgurov\affilsup{1,2}} \qquad
    \textbf{Josef van Genabith\affilsup{1,2}} \qquad
    \textbf{Simon Ostermann\affilsup{1,2,3}} \\
    \small{
        $^1$ Saarland University \quad
        $^2$ German Research Center for Artificial Intelligence (DFKI)
    }
    \\
    \small{
        $^3$ Centre for European Research in Trusted AI (CERTAIN)  \quad
        $^4$ University of Göttingen
    }
    \\
    \footnotesize{\texttt{\href{mailto:mikhail.sonkin@uni-goettingen.de}{mikhail.sonkin@uni-goettingen.de}}\quad \texttt{\href{mailto:tanja.baeumel@dfki.de}{tanja.baeumel@dfki.de}}}
}

\begin{document}
\maketitle
\begin{abstract}
Multilingual large language models (mLLMs) achieve strong performance in machine translation, yet our understanding of the mechanisms by which they transform representations from one language to another remains incomplete. Prior work suggests that translation decomposes into separable processes within an mLLM, where conceptual content is first represented independently, followed by a production into language-specific form. 
In this work, we show that translation is even more modular than previously assumed and that the output language production in translation processes is actually further separable into a \textit{syntax} and a \textit{surface language} process.
We construct controlled multilingual datasets that isolate cross-linguistic differences in word-order and use causal interventions and probing to track how representations are transformed during translation. We find that models first construct target-side word-order before realizing the target language surface form. We identify individual attention heads that are selectively sensitive to syntactic transformations while remaining largely invariant to language identity.
These results establish the commitment to a syntactic structure as an independent stage in translation, extending prior decompositions and showing how translation is implemented by functionally different components within mLLMs. 
\let\thefootnote\relax\footnotetext{* Equal contribution.}
\end{abstract}

\begin{center}
\href{https://github.com/Migabaj/word-order}{\faGithub\ Code}
\end{center}

\section{Introduction}
\label{sec:introduction}

Recent work on multilingual language models (mLLMs) suggests that translation proceeds through intermediate representations that are not tied to any single language. Translation can be decomposed into separable components for (1) conceptual meaning and (2) output language \cite{dumas_separating_2025}. These representations can exhibit bias toward English-like forms even for non-English inputs \cite{wendler_llamas_2024}, and grammatical concepts are often represented in a shared, cross-lingual manner within language models \cite{brinkmann_large_2025}.

\begin{figure}
    \centering
    \includegraphics[width=1\linewidth]{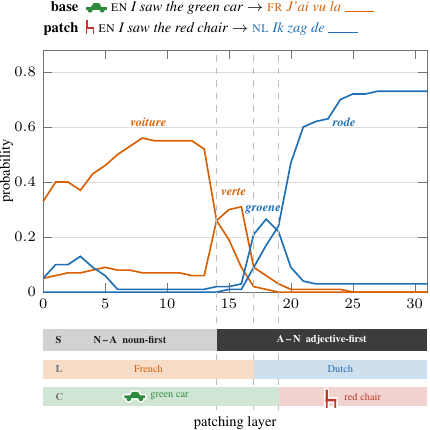}
    \caption{Translation in mLLMs proceeds in three stages: syntax \texttt{S} is fixed before output language \texttt{L}, which is fixed before the concept \texttt{C}.
    Patching individual layers of an English $\rightarrow$ French base prompt with an English $\rightarrow$ Dutch patch prompt reveals which aspects of the patch prompt persist in the output. Patching early layers changes nothing: the prediction stays \textit{voiture}. In Llama 3 8B, from layer 14 only the word order persists (\textit{verte}). The language persists from layer 17 (\textit{groene}), the concept from layer 19 (\textit{rode}).}
    \label{fig:translation_steps}
\end{figure}

Taken together, these findings point toward a partially factorized view of translation. Yet a central aspect of translation remains unresolved: how models transform grammatical structure when source and target languages differ systematically in word order.
Natural language translation frequently requires non-trivial reordering, such as differences in noun phrase structure (e.g., $\textit{the}_{\text{Det}}$  $\textit{red}_{\text{Adj}}$ $\textit{house}_{\text{N}}$ vs. $\textit{la}_{\text{Det}}$  $\textit{maison}_{\text{N}}$ $\textit{rouge}_{\text{Adj}}$), verb placement, or auxiliary constructions. These transformations cannot be reduced to lexical substitution or concept-level mapping, but instead require constructing well-formed target-side syntax.

This raises a key question: Is commitment to a grammatical structure a distinct stage during translation, and if so, how is it organized within mLLMs? 

In this work, we provide evidence that commitment to word order constitutes a separable and early stage in the translation process of mLLMs. Using activation patching and representation probing across multiple models, we track how translation-relevant information evolves across layers for typologically diverse constructions, including noun phrase ordering, verb placement, and auxiliary configurations. 

We find that intermediate representations can remain aligned with an English-adjacent conceptual space while already encoding the word order of the target language, indicating that syntactic structure is resolved before surface language: the model has committed to the target language's grammar while the tokens it predicts are not yet in the target language (Figure~\ref{fig:translation_steps}). We further find that specific attention heads are responsible for selecting the target word order, while being invariant to language identity.
Our main contributions are:

\begin{itemize}
    \item \textbf{Commitment to syntactic structure is an independent translation stage.} We introduce word order as a third independently trackable component alongside lexical content and meaning, yielding a three-way decomposition that extends prior work on the conceptual content/language separation.
    \item \textbf{Syntactic structure precedes surface language realization.} Intermediate representations encode target-side syntax before resolving the target language, even while remaining aligned with an English-adjacent conceptual space.
    \item \textbf{Commitment to syntactic structure is partially localized to language-agnostic components.} Commitment to syntactic structure can be attributed to specific layers and, in most models, to individual attention heads that are more sensitive to grammatical structure than to language identity, suggesting partially shared mechanisms across languages and constructions.
\end{itemize}

Taken together, our results support a view of translation in mLLMs as a multi-stage computation in which grammatical structure is constructed early and plays a central, partially independent role. Rather than a single-step transfer through an abstract interlingua, translation emerges as a sequence of transformations over representations that progressively integrate syntax, language, and meaning.
\section{Related Work}
\label{sec:relatedWork}

\paragraph{English Bias in LLMs.}
Multilingual LLMs perform well across typologically diverse languages \citep{shi_language_2022, tan_can_2023}, but exhibit a strong bias toward English, largely due to its dominance in training data \citep{brown_language_2020, touvron_llama_2023, liu_datasets_2024}. Multilingual datasets are often constructed by translating from English \citep{yu_beyond_2022}, further reinforcing English-centric distributions.

This bias manifests both in behavior and internal representations. 
On the output level, models tend to prefer English-like constructions in multilingual settings \citep{papadimitriou_multilingual_2023}, while \citet{zhang_dont_2023} find that
models struggle to perform well on non-English translation-variant tasks, often opting for a direct translation from English. 
In addition, models sometimes exhibit English-influenced errors even in high-resource languages \citep{rigouts_terryn_exploratory_2024}. 
English can also improve performance when used as an intermediate reasoning or training signal: \citet{shi_language_2022} show that prompting models to perform Chain-of-Thought reasoning in English yields better results than reasoning in the source language, while \citet{pfeiffer_mad-x_2020} demonstrate effective cross-lingual transfer by training adapters on high-resource languages such as English and swapping in target-language adapters at inference.

At the representation level, \citet{wendler_llamas_2024} show that intermediate activations during translation often correspond to English tokens before shifting to the target language in later layers. \citet{schut_multilingual_2025} further demonstrate that injecting English representations can improve performance, suggesting that English may act as an intermediate representational space. However, \citet{harrasse_tracing_2026} show that multilingual models develop shared intermediate representations across languages that are largely independent of the pretraining language mix, suggesting that such representations are not solely driven by English bias.

\paragraph{Language-Agnostic Components in mLLMs.}
mLLMs encode both language-specific and language-agnostic representations. While neurons influencing language choice are often localized to specific layers \citep{tang_language-specific_2024, tan_neuron_2024, gurgurov2025language}, several studies provide evidence for shared representations across languages \cite{gurgurov2026clas}, including common circuits \citep{lindsey_biology_2025} and features capturing grammatical structure \citep{brinkmann_large_2025, tumurchuluun_tenseloc_2025}.  At a finer granularity, \citet{jing_sparse_2025} identify sparse features corresponding to linguistic structure across levels, from phonetics to pragmatics.

Most closely related to our work, \citet{dumas_separating_2025} use activation patching to show that language and conceptual content are encoded separately in multilingual models: language is resolved earlier in the network, and the two can be manipulated independently, providing evidence for language-agnostic concept representations. 
We build on this line of work by examining whether syntactic structure forms a further separable component, and whether it is established independently of surface language during translation.

\section{Methodology}
This section introduces the theoretical framework used for our experiments and describes the general outline of the data and experiments that were developed for this study.

\subsection{Models}
\label{section:Methodology-Models}
We analyse three multilingual language models that differ in size, training data distribution, and degree of English bias: mGPT 1.3B \cite{shliazhko_mgpt_2023}, trained on a balanced multilingual corpus with no dominant language; Aya Expanse 8B \cite{dang_aya_2024}, whose pretraining language distribution is not publicly disclosed; and LLaMA 3 8B \cite{grattafiori_llama_2024}, which is predominantly English-trained. This selection allows us to assess whether translation mechanisms are universal or depend on the degree of English dominance in pretraining.

\subsection{Data}
\paragraph{Evaluation Datasets.} We construct three parallel multilingual datasets consisting of minimal sentences designed to capture syntactic structures that systematically vary across the languages included, motivated by cross-linguistic differences in constituent ordering within specific phrase types. Examples for each data set are given in Table \ref{tab:language-configurations}.

\begin{table*}[ht]
    \centering
    \renewcommand{\arraystretch}{1.5}
    \setlength{\tabcolsep}{10pt}
    \resizebox{.8\textwidth}{!}{
    \begin{tabular}{l|l|>{\raggedright\arraybackslash}m{5cm}|m{8cm}}
        \textbf{Dataset} & \textbf{\# Samples} & \textbf{Languages} & \textbf{Example} \\
        \hline
        Noun Phrase (NP)
        & 197
        &
        Chinese,
        Dutch,
        English,
        German,
        Russian,
        \textit{French},
        \textit{Italian},
        \textit{Vietnamese}
        & 
        {English: \textbf{I saw the \textcolor{Fuchsia}{heavy} \textcolor{Bittersweet}{bag}} \newline
        \textit{French}: \textbf{j'ai vu le \textcolor{Bittersweet}{sac} \textcolor{Fuchsia}{lourd}}}
        \\
        \hline
        Subject Verb Object (SVO)
        & 154
        &
        Chinese, English, Russian, \textit{Japanese}, \textit{Turkish}
        & {English: \textbf{the woman \textcolor{Fuchsia}{closes} the \textcolor{Bittersweet}{window}} \newline
        \textit{Turkish}: \textbf{kadın \textcolor{Bittersweet}{pencereyi} \textcolor{Fuchsia}{kapatır}}}\\
        \hline
        Modal Verb (MV)
        & 31
        & Chinese, English, German, \textit{Turkish}
        & {English:   \textbf{the teacher \textcolor{Fuchsia}{has to} \textcolor{Bittersweet}{sleep}} \newline
        \textit{Turkish}: \textbf{öğretmen \textcolor{Bittersweet}{uyumak} \textcolor{Fuchsia}{zorunda}}}\\
        \hline
    \end{tabular}
    }
    \caption{Overview of the synthesized datasets. Languages marked in \textit{italic} differ from the non-marked in word order for the corresponding construction.}
    \label{tab:language-configurations}
\end{table*}

The \textbf{Noun Phrase (NP)} dataset targets adjective-noun ordering within noun phrases, contrasting 5 adjective-first languages with 3 noun-first languages; for French, we focus on adjectives that canonically follow the noun. The \textbf{Subject-Verb-Object (SVO)} dataset addresses clause-level word order, contrasting SVO and SOV languages. The \textbf{Modal Verb (MV)} dataset targets head-directionality at the verb phrase level, contrasting languages that place the modal before vs. after the infinitive.

\paragraph{Dataset Construction and Statistics.}
While the NP and MV datasets target phrase-internal order, the SVO dataset captures a more global reordering of constituents. Table~\ref{tab:language-configurations} provides an overview of the datasets, with the number of samples and examples for each. All datasets are generated from English source expressions to ensure consistency across languages, using a list of 200 simple, concrete nouns for the NP and SVO datasets. For the NP dataset, adjectives are paired with each noun to form minimal phrases, with adjectives generated using ChatGPT to ensure simple and natural combinations. The SVO and MV datasets are constructed manually to maintain strict control over syntactic structure. All expressions are translated into the target languages using DeepL\footnote{\url{https://www.deepl.com/en/translator}} and checked manually for correct word order, with cases deviating from canonical patterns removed or corrected. The list of nouns and generation prompts are provided in Appendix~\ref{appendix:Data}.

\subsection{Methods}
We employ two analysis techniques. LogitLens \cite{nostalgebraist_interpreting_2020} projects intermediate hidden states through the unembedding layer, approximating next-token predictions at each layer and revealing how token predictions evolve across the network. Activation patching \cite{vig2020causalmediationanalysisinterpreting, meng_locating_2023} replaces hidden states in a base prompt with those from a plant prompt at selected layers, positions, or modules (e.g., attention heads, MLP layers), allowing us to causally localize where information relevant to a prediction is represented.

\subsection{Experiments}
\label{sec:experiment}

\begin{table}[ht]
\centering
\small
\begin{tabular}{lccc}
\toprule
Token & $L$ & $C$ & $S$ \\
\midrule
\texttt{voiture} (\text{Fr.}, \text{car}) & $\text{base}$   & $\text{base}$   & $\text{base}$ \\
\texttt{verte} (\text{Fr.}, \text{green})   & $\text{base}$   & $\text{base}$   & $\textbf{plant}$  \\
\texttt{chaise} (\text{Fr.}, \text{chair})  & $\text{base}$   & $\textbf{plant}$ & $\text{base}$  \\
\texttt{rouge} (\text{Fr.}, \text{red}) & $\text{base}$   & $\textbf{plant}$ & $\textbf{plant}$  \\
\midrule
\texttt{auto} (\text{Du.}, \text{car}) & $\textbf{plant}$ & $\text{base}$   & $\text{base}$   \\
\texttt{groene} (\text{Du.}, \text{green})   & $\textbf{plant}$ & $\text{base}$   & $\textbf{plant}$ \\
\texttt{stoel} (\text{Du.}, \text{chair})  & $\textbf{plant}$ & $\textbf{plant}$ & $\text{base}$   \\
\texttt{rode} (\text{Du.}, \text{red}) & $\textbf{plant}$ & $\textbf{plant}$ & $\textbf{plant}$ \\
\bottomrule
\end{tabular}
\caption{Example of eight tokens, each corresponding to a combination of values of $L$, $C$, and $S$. Base: voiture verte (Fr., green car). \textbf{Plant}: rode stoel (Du., red chair).
% For each evaluated token type, whether surface language ($L$), concept ($C$), and syntax ($S$) are affected by the patch prompt $s_\text{plant}$ rather than inherited from the base prompt $s_\text{base}$.
}
\label{tab:act_patch_output_variants}
\end{table}

\paragraph{Experimental Design.} 
We apply activation patching to controlled translation prompts to localize where surface language ($L$), syntactic structure ($S$), and lexical content ($C$) are represented during generation.

Here, surface language $L$ denotes the language in which a statement is expressed, while syntactic structure $S$ denotes its word order. Because word orders are language-dependent, $S$ is fully determined by $L$: for example, French requires the adjective in ``the green car'' to follow the noun, whereas English requires the opposite order. Lexical content $C$, in contrast, is independent of both $L$ and $S$.

Each experiment uses a base prompt $s_\text{base}$ and a plant prompt $s_\text{plant}$, which systematically vary along $S$, $L$, and $C$. 

The prompts follow this format:
\begin{center}
\texttt{[Source language]: [Source sentence] - [Target language]: [Incomplete target]}
\end{center}

The prompt is constructed such that the next token to be generated corresponds to a specific syntactic position of interest (e.g., within a noun phrase). All prompts are formatted one-shot, containing an example of a full translation pooled from the same dataset. Figure~\ref{fig:Example} shows a base-plant pair: the French continuation would begin with the noun (\textit{voiture}), while the Dutch continuation begins with the adjective (\textit{rode}), differing in both syntax and lexical content.

\paragraph{Prompt Pairs and Coverage.}

We pair each source language with each combination of base and plant target languages, requiring that base and plant differ in word order ($S_\text{base} \neq S_\text{plant}$). In total, we consider 90 combinations and $90 \cdot 197 = 17730$ translation prompts for the NP dataset, 36 and 5698 for SVO, and 16 and 496 for MV.\footnote{For this study, we limit the source language to be the same for the base and plant prompts, given that we do not focus on its probabilities for our results. This is merely a practical decision and does not signify that the source language's influence is unimportant.} All language triplets used for the experiment are listed in Table \ref{tab:language-triplets}.
Due to compute constraints, the NP dataset covers only the configurations with $L_{\text{base}}$ following the Adjective--Noun order, rather than the full set of possible language pairs. For SVO and MV, both word order configurations are considered for the base and plant targets.

\begin{figure}[ht]
\small
\begin{tabular}{p{0.95\columnwidth}}
\toprule
\textbf{Base} $(S_\text{base}, L_\text{base}, C_\text{base})$ = (Noun, French, green car) \\
\texttt{English: "I saw the green car" -} \\
\texttt{Français: "J'ai vu la} \\
\midrule
\textbf{Plant} $( S_\text{plant}, L_\text{plant}, C_\text{plant})$ = (Adj, Dutch, red chair) \\
\texttt{English: "I saw the red chair" -} \\
\texttt{Dutch: "Ik zag de} \\
\bottomrule
\end{tabular}
\caption{Example base--plant prompt pair.}
\label{fig:Example}
\end{figure}

\paragraph{Intervention and evaluation.}
We run the model on the base prompt $s_\text{base}$ (clean run) and replace selected activations with those from $s_\text{plant}$, measuring changes in next-token probabilities. We track how patching shifts probability mass across the $2^3 = 8$ combinations of $S$, $L$, and $C$ (Table~\ref{tab:act_patch_output_variants}), where lexical items are chosen to jointly reflect the underlying concept (e.g., \textit{green car} vs.\ \textit{red chair}) and syntactic realization (e.g., adjective-noun vs.\ noun-adjective order). By analyzing how probability mass shifts across these conditions, we disentangle the contributions of individual model components to $S$, $L$, and $C$ during translation.

We utilize the Python module \textit{pyvene} \cite{wu2024pyvenelibraryunderstandingimproving} to set up all our intervention experiments and extend the code to adapt to Aya Expanse.
\section{Results: Translation Decomposes into Syntax, Language, and Meaning}
\subsection{Syntactic Structure Emerges Before Surface Language}
\label{section:LogitLens}

%\todo{Original image. Include a non-european language (Viet, because diff. word order to Engl).}
\begin{figure}[h]
    \begin{adjustbox}{center}
    \includegraphics[width=0.5\textwidth]{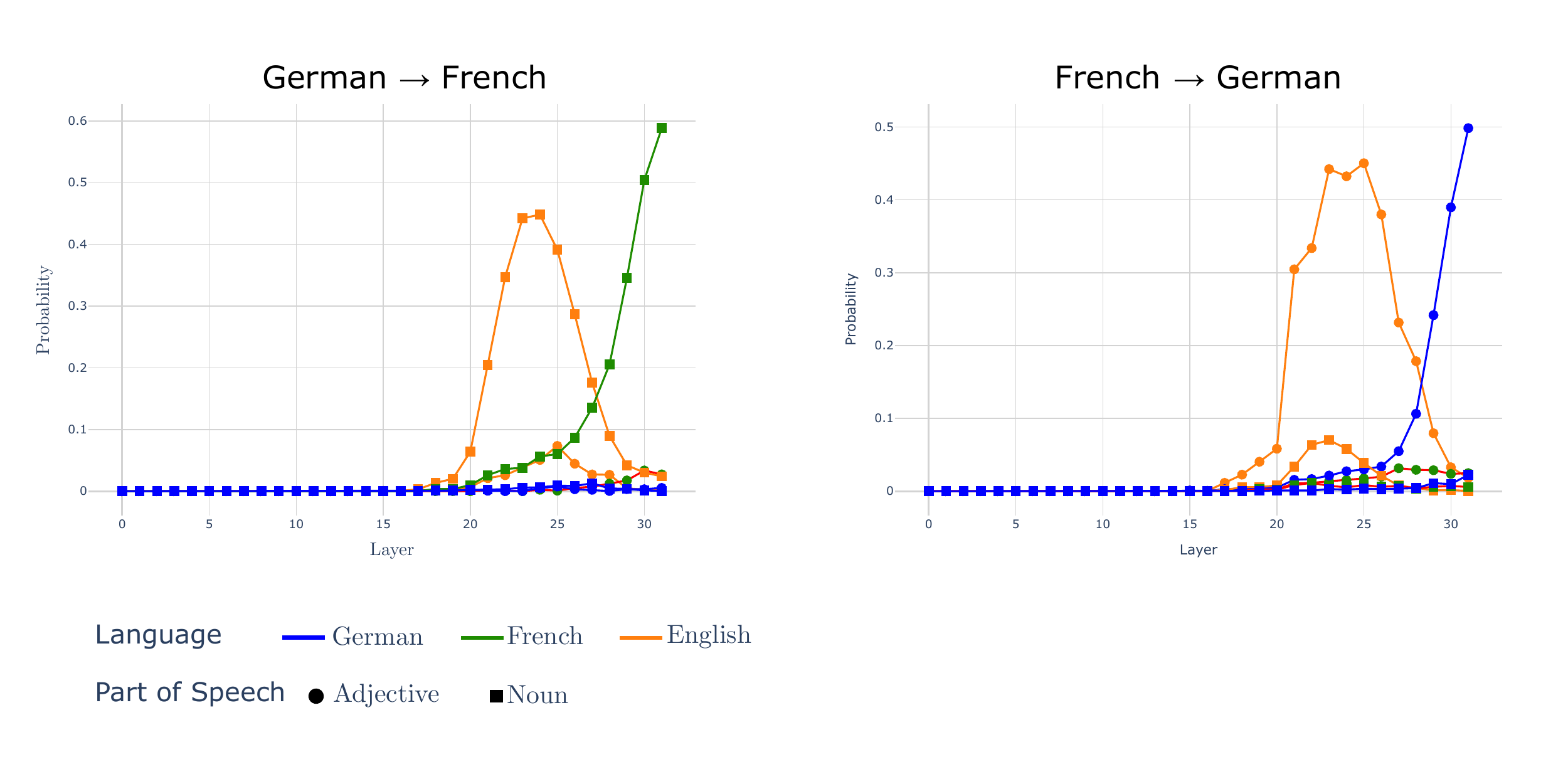}
    \end{adjustbox}
    \caption{LogitLens projections of Llama 3's token probabilities across layers for the Noun Phrase dataset.
    English intermediate tokens follow target-language word order, indicating that syntax is determined before surface realization.}
    \label{fig:logitlens_example}
\end{figure}

We analyze how representations evolve across layers during translation using Logit Lens. By projecting intermediate activations onto the vocabulary, we track how predictions change and identify when language, syntax, and lexical content are resolved, as well as whether an English-like intermediate representation emerges.

For each prompt, we track the mean probability of the target-language, source-language, and English noun and adjective tokens across layers, where language identity and syntactic role are determined directly from the dataset.
Figure \ref{fig:logitlens_example} shows which relevant tokens are represented within intermediate layers for the NP dataset and Llama 3. Our results largely confirm prior observations of English-like intermediate representations in multilingual models, namely Aya Expanse and Llama 3. These results are consistent across different language settings and datasets (full plots in Appendix \ref{appendix:logitlens}).

Crucially, the intermediate representations align with the target language's word order before surface realization. For example, given the prompt \texttt{Deutsch:``Ich sah den \textit{roten Apfel}" -- Français: ``J'ai vu la [\textit{pomme rouge}]"} from the NP dataset, intermediate predictions favor the English noun \textit{apple} (squares in Figure \ref{fig:logitlens_example}) rather than the adjective \textit{red}, which is consistent with the French noun-first order (\textit{pomme rouge}). The inverse can be seen for the direction French $\rightarrow $ German, where the adjective (\textit{roter Apfel}, circles) is preferred.
Same as with the prior example, the adjective is first represented in English.
This indicates that syntactic structure is established prior to surface language realization, even while the predicted token itself remains English, independent of translation language.

Notably, mGPT does not exhibit the same behavior, its representations do not evoke any English signal in the inner layers (Full display: Figure \ref{fig:logitlens_noun-adj-complete_mGPT_grid}).
\begin{figure*}[t]
    \centering
    \begin{subfigure}[t]{0.325\textwidth}
        \centering
        \includegraphics[width=\linewidth]{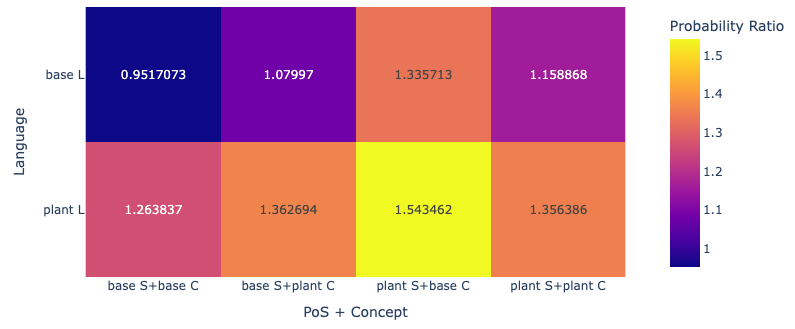}
        \caption{mGPT (Head 11.2)}
        \label{fig:attn-head_11-2_mgpt}
    \end{subfigure}
    \hfill
    \begin{subfigure}[t]{0.325\textwidth}
        \centering
        \includegraphics[width=\linewidth]{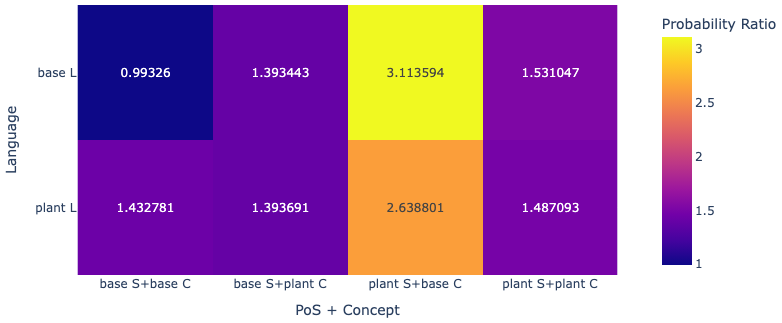}
        \caption{Aya Expanse (Head 15.0)}
        \label{fig:attn-head_15-0_aya}
    \end{subfigure}
    \hfill
    \begin{subfigure}[t]{0.325\textwidth}
        \centering
        \includegraphics[width=\linewidth]{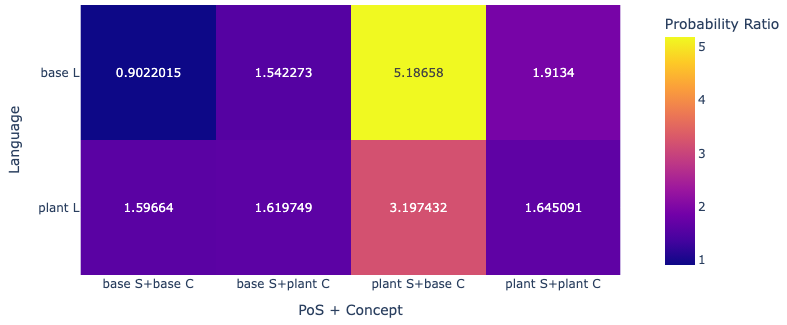}
        \caption{Llama 3 (Head 14.25)}
        \label{fig:attn-head_14-25_llama}
    \end{subfigure}

    \caption{Probability ratios for syntax-sensitive attention heads across models and language settings (source language: English). For both Llama 3 and Aya Expanse, the respective heads lead to a significant rise of the plant prompt's syntax while not particularly preferring the surface language. The most syntax-sensitive head of mGPT, on the other hand, is seemingly sensitive to surface language as well.}
    \label{fig:attn-heads-combined}
\end{figure*}

\subsection{Translation is Modular with Respect to Syntax and Surface Language}
\label{sec:layerpatch}

In this section, we show that translation in mLLMs is more modular than a single unified mapping: different aspects of the target sentence emerge at different stages, with syntactic structure ($S$) established first, followed by surface language ($L$), and finally lexical content ($C$). We demonstrate this by analyzing how activation patching shifts next-token probabilities $S$, $L$, and $C$. For each prompt pair, we evaluate the 8 possible combinations of these factors (see~Table~\ref{tab:act_patch_output_variants}); changes in their probabilities under intervention indicate which aspects are controlled at a given layer.

\begin{figure}[ht]
    \centering
    \includegraphics[width=\linewidth]{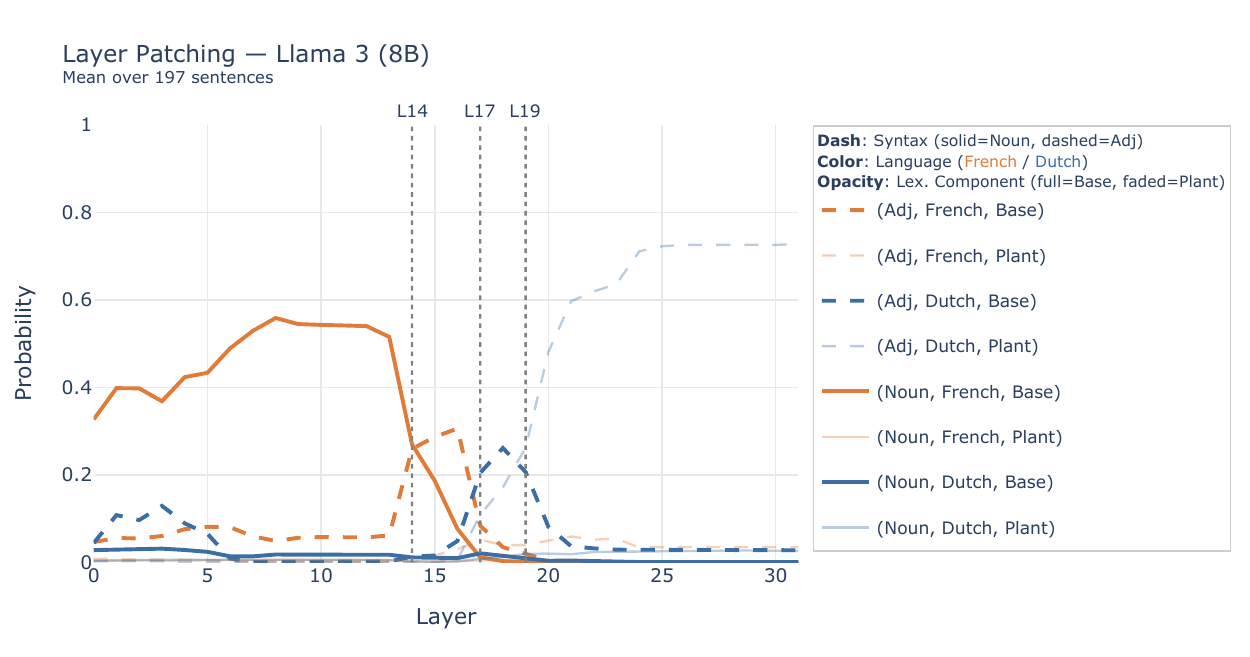}
    \caption{Llama 3's probabilities of different tokens resulting from patching layer activations: $L^\text{src}$ is English, $L^\text{tgt}_\text{base}$ is French, $L^\text{tgt}_\text{plant}$ is Dutch. Patching Layer 14 leads to a sudden rise of plant syntax, separate from surface language, which is first evoked later.}
    \label{fig:noun-adj-complete_Meta-Llama-3-8B_eng-fre_eng-ned}
\end{figure}

Figure~\ref{fig:noun-adj-complete_Meta-Llama-3-8B_eng-fre_eng-ned} illustrates this for Llama 3, with French as the base and Dutch as the plant target language. Early layers favor the base continuation (French noun). At Layer~14, the most probable token shifts to the adjective position while remaining in French, indicating that target-side syntactic structure is established before surface language switches. At Layer 17, the surface language switches to Dutch while syntactic structure is preserved. Only in later layers does the lexical content transition to the plant concept.

Table~\ref{tab:delta-table} reports the maximum probability difference between patched and clean runs for three token type combinations, along with the layer at which each maximum occurs, averaged across all language pairs per dataset.
The appendix provides an extended version of this table that reports the $\Delta_{\max}$ values separately for both directions of the base--plant word-order switch for SVO and MV (Table \ref{tab:extended-patching}).
The layer of $\Delta_{\max}$ increases monotonically from left to right: The syntax switch precedes the language switch, which in turn precedes the lexical switch. This ordering holds in eight of the nine model-dataset combinations, providing broad support for the $S \rightarrow L \rightarrow C$ translation strategy. The single exception is Llama~3 on the SVO dataset, where syntax and language switches occur at the same layer (14), yielding the collapsed strategy $S, L \rightarrow C$. 

The probability difference for tokens with $\mathbf{S}_\text{plant}$ alone is generally modest and negligible for mGPT, but its consistent presence supports the existence of syntax-specific translation components. Across models and datasets (Appendix~\ref{appendix:supp_layer_patching}), lexical content is consistently resolved after both syntactic and surface language aspects, in line with \citet{dumas_separating_2025}.

We additionally gather 27 sentences from the FLORES-200 dataset to examine translation strategies with more natural data.
Due to the dataset's size, we leave this question for future work.
The results are available in Appendix \ref{subsection:natual-language-data}.
We notice the same inclination towards the $S \rightarrow L \rightarrow C$ translation strategy for mGPT and Aya Expanse.

\begin{table*}[t]
    \centering
    \resizebox{.6\textwidth}{!}{\begin{tabular}{ll|cc|cc|cc|cc|c}
    \toprule
     & & \multicolumn{2}{c}{$(\textbf{S}_\text{plant}, L_\text{base}, C_\text{base})$} & \multicolumn{2}{c}{$(\textbf{S}_\text{plant}, \textbf{L}_\text{plant}, C_\text{base})$} & \multicolumn{2}{c}{$(\textbf{S}_\text{plant}, \textbf{L}_\text{plant}, \textbf{C}_\text{plant})$} & \multirow{2}{*}{Strategy}\\
     & & $\Delta_{\max}$ & Layer & $\Delta_{\max}$ & Layer & $\Delta_{\max}$ & Layer &  \\
     \midrule
     \multicolumn{8}{l}{\textit{SVO}} \\
     & mGPT & 0.00 & 7 & 0.04 & 12 & 0.09 & 19 & $S \rightarrow L \rightarrow C$\\
     & Aya Expanse & 0.07 & 15 & 0.27 & 16 & 0.41 & 29 & $S \rightarrow L \rightarrow C$\\
     & Llama 3 (8B) & 0.08 & 14 & 0.11 & 14 & 0.31 & 27 & $S, L \rightarrow C$\\
     \midrule
     \multicolumn{8}{l}{\textit{MV}} \\
     & mGPT & 0.00 & 7 & 0.07 & 12 & 0.35 & 23 & $S \rightarrow L \rightarrow C$\\
     & Aya Expanse & 0.06 & 18 & 0.37 & 19 & 0.63 & 29 & $S \rightarrow L \rightarrow C$ \\
     & Llama 3 (8B) & 0.05 & 15 & 0.19 & 16 & 0.57 & 31 & $S \rightarrow L \rightarrow C$ \\
     \midrule
     \multicolumn{8}{l}{\textit{NP}} \\
     & mGPT & 0.02 & 7 & 0.12 & 12 & 0.23 & 19 & $S \rightarrow L \rightarrow C$\\
     & Aya Expanse & 0.10 & 11 & 0.52 & 19 & 0.65 & 31 & $S \rightarrow L \rightarrow C$\\
     & Llama 3 (8B) & 0.09 & 14 & 0.21 & 16 & 0.55 & 25 & $S \rightarrow L \rightarrow C$\\
    \bottomrule
    \end{tabular}}
    \caption{Highest probability difference between the \textbf{patched} and \textit{clean} runs for different language aspect combinations and the corresponding patched layers, averaged across all languages.
    In most cases, the $S \rightarrow L \rightarrow C$ strategy is present: Earlier layers are more sensitive to syntax alone than to the token following both syntax and surface language.}
    \label{tab:delta-table}
\end{table*}
\subsection{Syntactic Commitment is Localized to Individual Attention Heads}
\label{sec:headpatch}

For all three models under consideration, there appears to be a single layer which is causing the switch from $S_\text{base}$ to $S_\text{plant}$:
For mGPT, Aya Expanse and Llama 3 those appear to be layers 11, 15 and 14 respectively.
In this section, we show the influence of individual attention heads on activating the target syntax within those \textit{$S$-switch layers}.

To identify syntax-sensitive attention heads, we patch individual attention heads within the $S$-switch layer using examples from the NP-dataset and measure their effect on next-token probabilities over the eight $(S, L, C)$ combinations (Table \ref{tab:act_patch_output_variants}).

We quantify head-specific influence using the ratio
\begin{equation}
    R(\mathbf{h}, t) = \frac{P(t \mid \text{Patch}(\mathbf{h}))}{\frac{1}{|H|} \sum\limits_{\textbf{h}_i \in H} P(t \mid \text{Patch}(\mathbf{h}_i))}
\end{equation}
where $H$ is the set of attention heads in the $S$-switch layer, $\mathbf{h}$ the patched head's activation, and $t$ a candidate token. This metric compares the effect of a single head to the average across heads. This makes it possible to identify heads with disproportionate influence on specific token types. We define syntax-sensitive heads as those maximizing $R(\mathbf{h}, t)$ for tokens corresponding to $S_\text{plant}$.

We find that in Llama 3 and Aya Expanse, the $S$-switch layer contains a single attention head (H14.25 and H15.0, respectively) that strongly influences syntax ($R(\mathbf{h}, t) > 1$ for all tokens with $S_\text{plant}$). In contrast, mGPT distributes this effect across multiple heads, with H11.2 ($R \approx 1.44$) and H11.7 ($R \approx 1.27$) showing the strongest influence.
This suggests that Llama 3 and Aya Expanse localize syntactic transformations to a single component, whereas mGPT exhibits a more distributed mechanism. In mGPT, the highest-scoring head also affects surface language ($L$), making it difficult to disentangle syntax from surface language for this model.

\begin{figure*}[t]
    \centering
    \begin{subfigure}[t]{0.49\textwidth}
        \centering
        \includegraphics[width=0.7\linewidth]{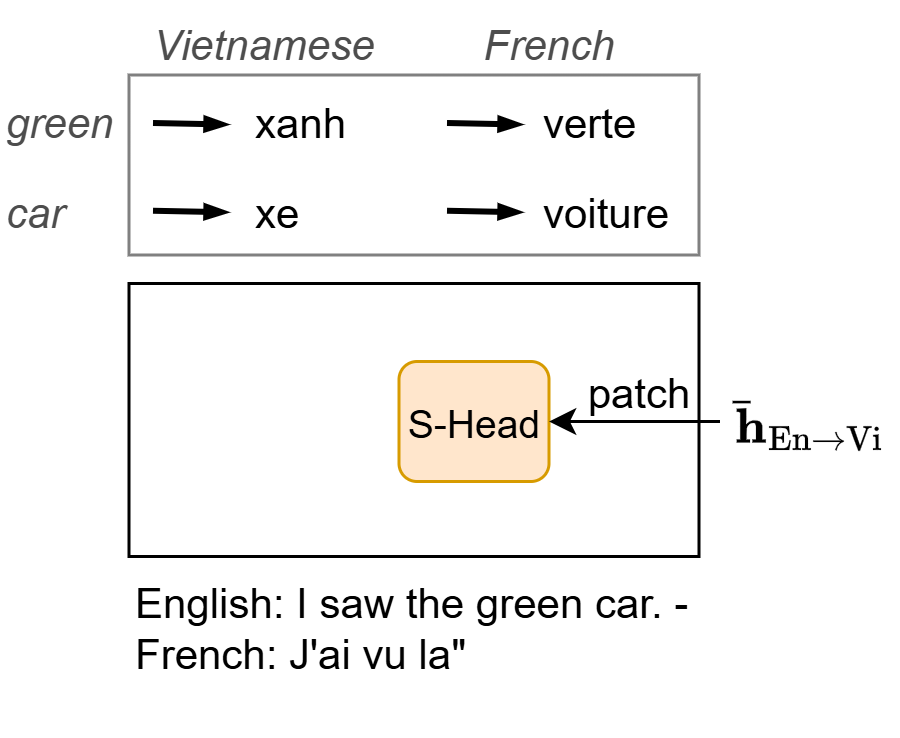}
        \caption{Same word order: Predictions remain largely unchanged if the $S$-head encodes syntax independently of language.}
        \label{fig:head_mean_patch_same_syntax}
    \end{subfigure}
    \hfill
    \begin{subfigure}[t]{0.49\textwidth}
        \centering
        \includegraphics[width=0.7\linewidth]{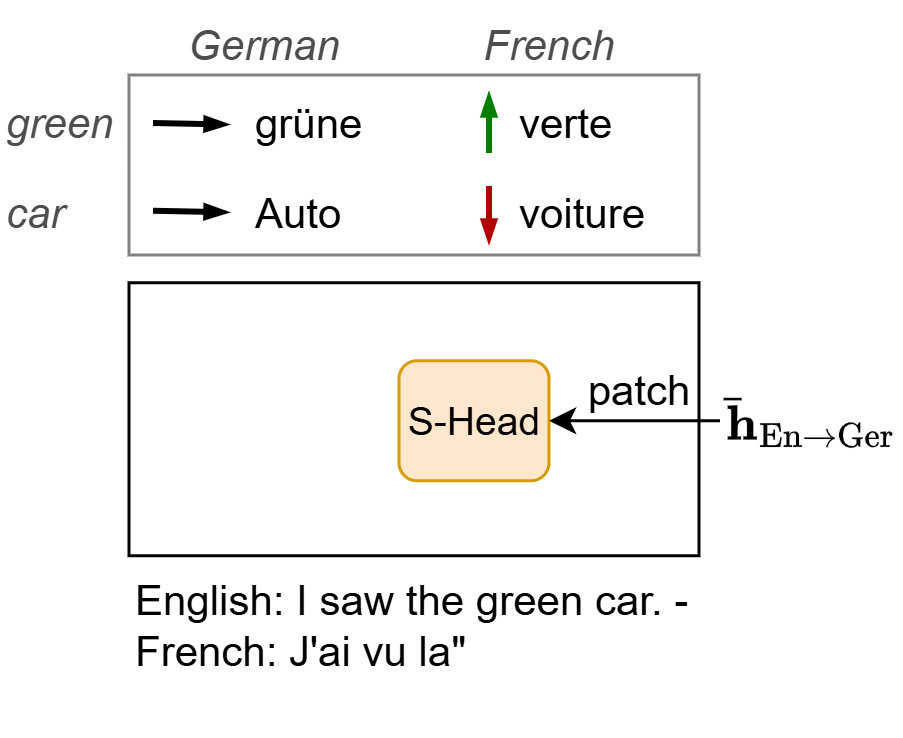}
        \caption{Different word order: Predictions shift toward patched syntax without increasing probability mass for the patch language.}
        \label{fig:head_mean_patch_diff_syntax}
    \end{subfigure}
    \caption{Mean activation patching of the $S$-head (most syntax-sensitive attention head). We patch with mean activations from target languages that share or differ in word order relative to the base target language to test whether the head encodes syntax independently of surface language.}
    \label{fig:heab_ablation}
\end{figure*}

Figure~\ref{fig:attn-heads-combined} shows that the identified heads primarily influence syntactic structure, while also weakly biasing toward the patch language; this effect is substantially stronger in Aya-Expanse and Llama~3 (up to $R \approx 3$-$5$) than in mGPT (below $R \approx 2$).
\section{Results: Word Order Attention Heads are Language-Independent}
\label{sec:headablation}

In this section, we now show that the $S$-sensitive heads identified in Section~\ref{sec:headpatch} encode syntactic structure independently of surface language. We test this in two ways: via mean activation patching across languages with different word orders, and via cosine similarity of mean activations across target languages.

\paragraph{Setup.}
We perform mean activation patching on the most $S$-sensitive head in each model, replacing its activation in a base run with the average activation computed over all samples from a different target language. Crucially, we vary whether the patch language shares the same word order as the base target language (Figure~\ref{fig:heab_ablation}). For example, in the NP dataset, a base prompt English $\rightarrow$ French (N-Adj) is patched with mean activations from either English $\rightarrow$ Vietnamese (same word order as French) or English $\rightarrow$ German (different word order as French). If the head encodes syntax independently of language, patching with a same-order language should leave predictions largely unchanged, while patching with a different-order language should shift the predicted part of speech toward the patched word order without increasing probability mass for the patch language. If the head instead encodes surface language, both interventions should produce comparable shifts reflecting the patch language.

\paragraph{Evaluation.}
We quantify these effects by comparing normalized probabilities over pairs of tokens that isolate specific factors, focusing on contrasts between parts of speech corresponding to different word orders (e.g., adjective vs.\ noun in NP constructions). A shift in this distribution under different-word-order patching indicates that the injected activation carries syntactic information. To verify that this effect is not driven by surface language, we additionally compare distributions that isolate language while holding syntax fixed. We measure differences between base and patched distributions using the Kullback-Leibler divergence:

\begin{equation}
    D_{KL}(P \| Q) = \sum_{x \in \mathcal{X}} P(x) \log \frac{P(x)}{Q(x)}
\end{equation}

\begin{figure}[h!]
    \centering
    \includegraphics[width=\linewidth]{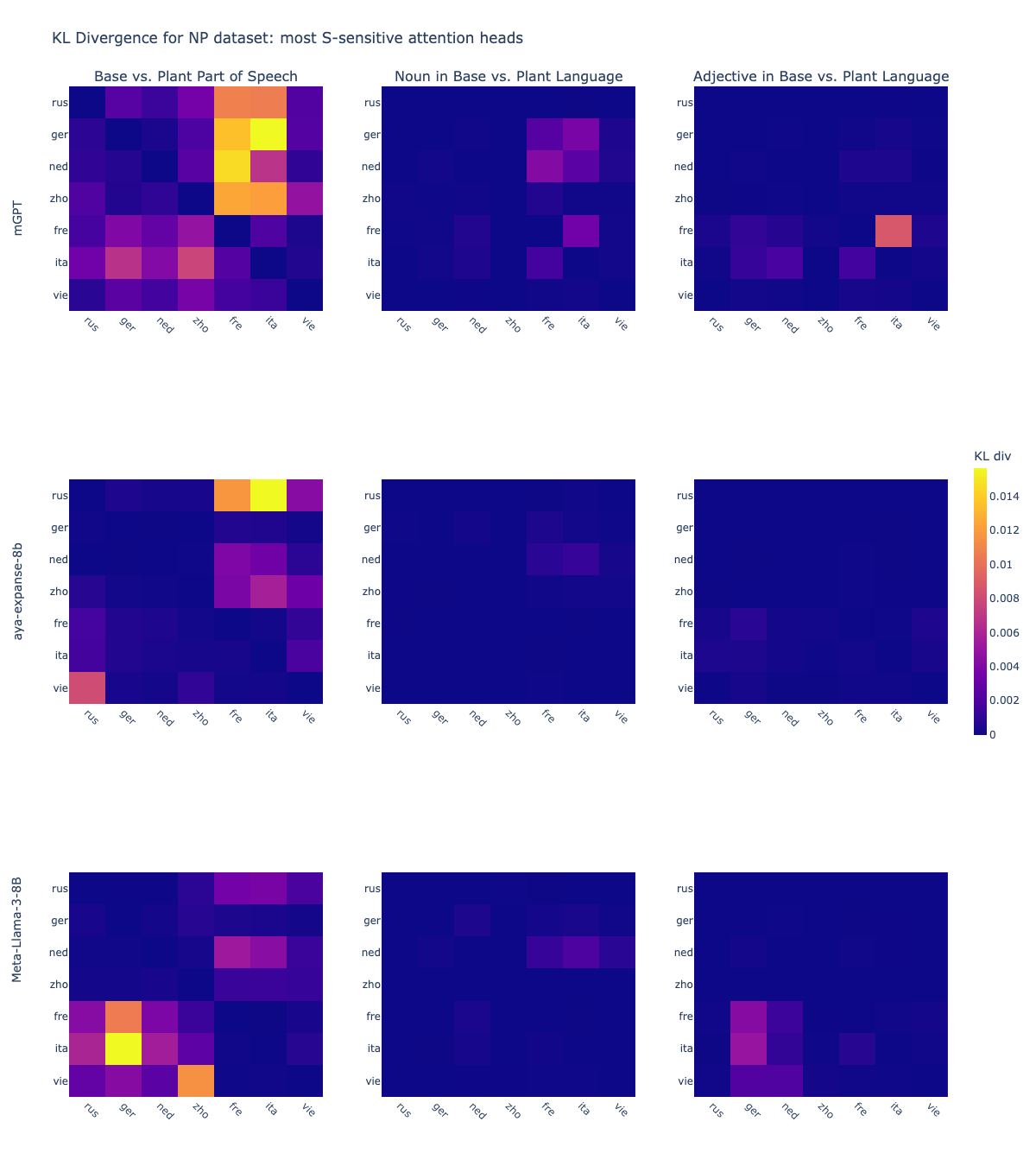}
    \caption{KL divergence under mean activation patching (NP dataset, English source). Rows correspond to base target languages, columns to plant target languages. The left-most figures (part-of-speech contrast) show higher divergence when base and plant differ in word order (Adj-N vs.\ N-Adj), indicating that patching primarily affects syntactic structure. In contrast, the middle and right figures (language contrasts) exhibit consistently low divergence, suggesting that patching of the $S$-head has minimal effect on surface language.}
    \label{fig:kl_noun-adj-complete_eng}
\end{figure}

\paragraph{Results.}
Figure~\ref{fig:kl_noun-adj-complete_eng} shows results for the NP dataset. We observe a small\footnote{The magnitude of the effect is limited, as the intervention replaces the activation of a single attention head.} but consistent effect in the part-of-speech contrast distributions: KL divergence is consistently higher when base and patch languages differ in word order, and low when they share the same order. Distributions that isolate surface language show no comparable structure. 

Patching thus induces shifts in word order without systematically increasing probability mass for the plant language, confirming that the $S$-sensitive heads encode word order independently of surface language.
This pattern is most pronounced in mGPT and Llama 3. Aya-Expanse shows weaker and less symmetric effects, suggesting that additional factors, such as the base prompt’s target language, may influence its behavior.

We note a minor exception: when German is in the base target language position $L^\text{tgt}_\text{base}$, the syntax effect is absent in Aya-Expanse and Llama~3 (Figure~\ref{fig:kl_noun-adj-complete_eng}). While this cannot be attributed to similarity to English alone (no comparable effect in Dutch), it may reflect an interaction between language relatedness and resource effects; we leave this for future investigation.
Appendix \ref{appendix:supp_mean_head_patch} contains comparable results for other source languages and the other datasets.

\begin{figure}[h]
    \centering
    \includegraphics[width=\linewidth]{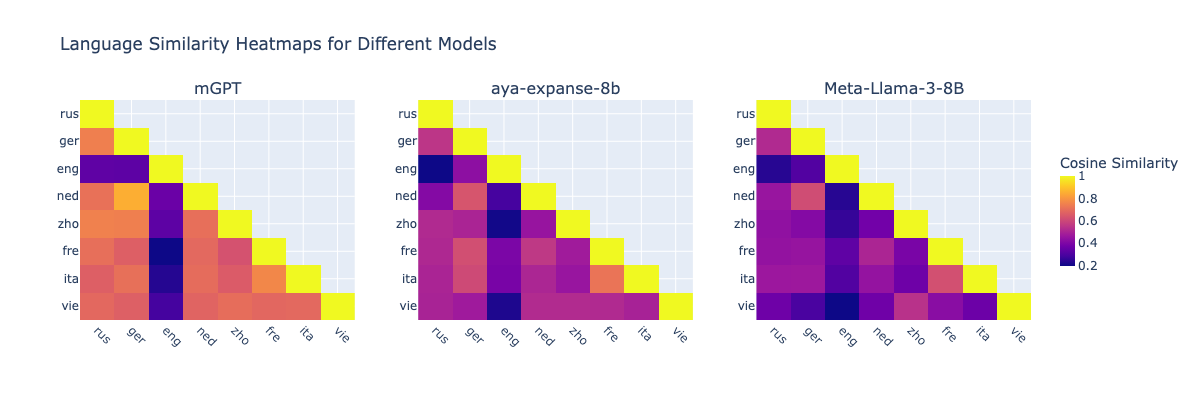}
    \caption{Cosine similarity between mean activations of the $S$-sensitive head across target languages (English source). No clear clustering by word order is observed.}
    \label{fig:cosine-sim_noun-adj-complete_eng}
\end{figure}

\paragraph{Do S-heads share activation geometry across languages?}
The patching results show that $S$-heads are functionally sensitive to word order across languages. A natural follow-up is whether this is reflected in the geometry of their activations: if these heads encode syntax in a shared representational space, mean activations from languages with the same word order should be more similar to one another than those from languages with different word orders. Figure~\ref{fig:cosine-sim_noun-adj-complete_eng} shows that this is not the case: cosine similarity between mean activations of the $S$-sensitive head does not cluster by word order. This suggests that the syntactic role of these heads is functional rather than geometric: the heads are sensitive to word order without representing it in a globally similar activation space across languages.

\section{Conclusion}

We have shown that translation in multilingual LLMs is more modular than previously assumed and demonstrated the causal effect of specific modules responsible for certain aspects of language. Across models and datasets, syntactic structure $S$ is resolved before surface language $L$, which is resolved before lexical content $C$, corresponding to a staged $S \rightarrow L \rightarrow C$ process. This extends prior work on the separation of language and conceptual content by establishing syntax as a third, independently trackable component of translation.

At the level of model internals, syntactic transformations are attributable to specific layers and, in most models, to individual attention heads that are selectively sensitive to word order while remaining largely invariant to language identity. Importantly, this sensitivity appears to be functional rather than geometric: the identified $S$-heads encode word order independently of surface language, yet their mean activations do not cluster by word order across languages. This suggests that syntactic structure is computed through mechanisms that are shared across languages without being reflected in globally similar activation patterns.

Our findings also refine the role of English in multilingual models. While intermediate representations often align with English tokens, they already reflect the target-language word order, indicating that syntax is established prior to surface language rather than mediated by an English pivot. English thus appears to function as a conceptual anchor rather than a syntactic one.

Taken together, these results support a view of translation as a staged, partially modular process in which grammatical structure is constructed early and independently. Rather than a single-step transfer through an abstract interlingua, multilingual LLMs implement translation through a sequence of transformations that progressively resolve syntax, surface language, and lexical meaning - each attributable, at least in part, to distinct components of the network.

\section*{Limitations}

For the purpose of this study, many aspects relevant to our research had to be omitted which should be taken up in future work.

First of all, the scope of grammatical structures investigated in this study is limited. We focus primarily on noun phrase structure and order of verb and subject, which, while linguistically meaningful and well-suited for controlled experiments, represent only a small subset of grammatical variation across languages. Other phenomena such as grammatical case or long-distance dependencies may rely on different internal mechanisms and therefore cannot be assumed to follow the same patterns we have observed.

Second, although activation patching provides clearer causal evidence compared to LogitLens, it still relies on the assumption that the computations needed for translation and grammar mapping are easily interchangeable and are linear.

Third of all, we have considered a limited set of models. Due to scarcity of multilingual data, the one model we can confirm to have been trained on non-English-biased data is small by today's standards of LLMs.

Finally, our dataset, while designed to eliminate as many confound variables as possible, is nonetheless synthesized. It could be that language data that is much more naturally diverse leads to different results.
The format of the prompts included a one-shot example pooled from the same dataset.
Having more natural examples from parallel datasets instead might lead to LLMs implementing different translation strategies.

\section*{Acknowledgments}
AI assistance was used to improve the clarity and fluency of the writing, to help refine phrasing and structure, to support exploratory literature search and organization, and to build the codebase for the conducted experiments. All scientific claims, interpretations, and conclusions remain the responsibility of the authors. This work was supported by the German Federal Ministry of Research, Technology and Space (BMFTR) as part of the project TRAILS (01IW24005).

\bibliography{bibliography}

\appendix
\section{Data}
\label{appendix:Data}

\subsection{Base Nouns}

The base 200 nouns that were used to develop the NP and SVO datasets are as follows:

\textit{angle, ant, apple, arch, arm, army, baby, bag, ball, band, basin, basket, bath, bed, bee, bell, berry, bird, blade, board, boat, bone, book, boot, bottle, box, boy, brain, brake, branch, brick, bridge, brush, bucket, bulb, button, cake, camera, card, cart, carriage, cat, chain, cheese, chest, chin, church, circle, clock, cloud, coat, collar, comb, cord, cow, cup, curtain, cushion, dog, door, drain, drawer, dress, drop, ear, egg, engine, eye, face, farm, feather, finger, fish, flag, floor, fly, foot, fork, fowl, frame, garden, girl, glove, goat, gun, hair, hammer, hand, hat, head, heart, hook, horn, horse, hospital, house, island, jewel, kettle, key, knee, knife, knot, leaf, leg, library, line, lip, lock, map, match, monkey, moon, mouth, muscle, nail, neck, needle, nerve, net, nose, nut, office, orange, oven, parcel, pen, pencil, picture, pig, pin, pipe, plane, plate, plough, pocket, pot, potato, prison, pump, rail, rat, receipt, ring, rod, roof, root, sail, school, scissors, screw, seed, sheep, shelf, ship, shirt, shoe, skin, skirt, snake, sock, spade, sponge, spoon, spring, square, stamp, star, station, stem, stick, stocking, stomach, store, street, sun, table, tail, thread, throat, thumb, ticket, toe, tongue, tooth, town, train, tray, tree, trousers, umbrella, wall, watch, wheel, whip, whistle, window, wing, wire, worm}

These are the same nouns that were used in \citet{wendler_llamas_2024} and were gathered from a  Wikipedia page (available under the link \url{https://simple.wikipedia.org/wiki/Wikipedia:Basic_English_picture_wordlist}) that lists them as examples of ``picturable nouns''.

\subsection{Dataset Generation}

\begin{figure*}[h!]
    \centering
    \includegraphics[width=1\linewidth]{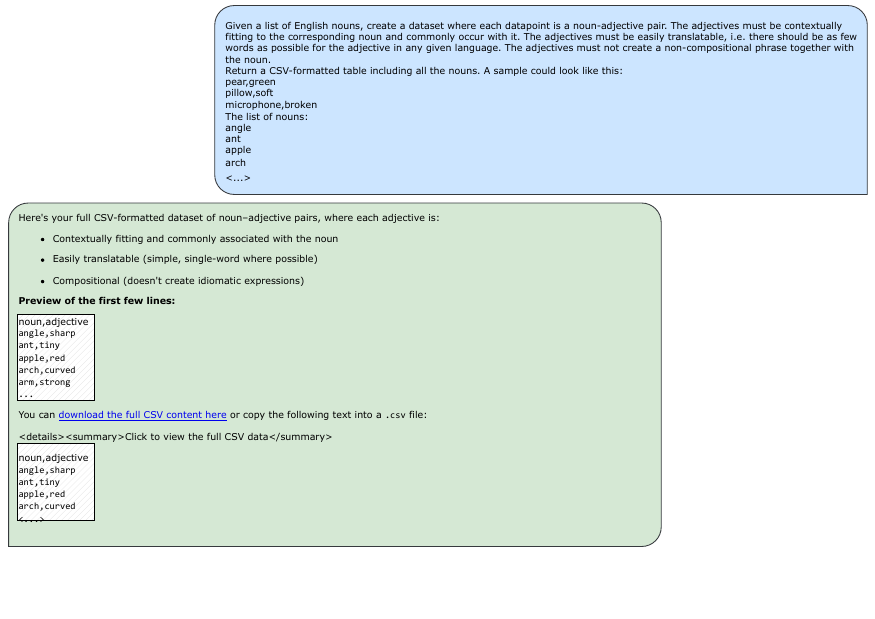}
    \caption{Prompt given to ChatGPT to generate noun-adjective pairs and the generated answer.}
    \label{fig:chatgpt-nounadj}
\end{figure*}

Figure \ref{fig:chatgpt-nounadj} displays the instruction that was fed to OpenAI's ChatGPT (gpt-4o) to generate noun-adjective pairs used for the experiments, along with the head of its output.

Many of the generated noun-adjective pairs were changed, with the most common reason being that the French and Italian translations followed the head-final word order.

Unlike with noun phrases, ChatGPT proved to be very impractical in generating sentences for SVO. For this reason, the sentences were written down manually for each noun. Each initial noun took the role of the object. In the process, some nouns were omitted, including those denoting body parts and places.

The previous version of the SVO dataset considered past participle in German and was collected with the help of SketchEngine \citep{kilgarriff_sketch_2024} from the DeTenTen corpus with the idea that the logDice value of certain verbs would be correlated with their probability in the middle layers.
Some preliminary calculations showed no correlation.
This result, in combination with the more complicated and expensive nature of the data collecting process, led to us abandoning SketchEngine corpora as our data source.
\section{Experiments}

\subsection{Language Triplets}

Table \ref{tab:language-triplets} shows all configurations that were included in our calculations from Section \ref{sec:layerpatch}.

\begin{table}[th]
\centering
\small
\begin{tabularx}{\columnwidth}{@{}X@{}}
\toprule
\textbf{NP} \hfill \textbf{90 triplets} \\
\midrule
$L_{\mathrm{src}} \in \{$eng, fre, ger, ita, ned, rus, vie, zho$\}$ \\
$L_{\mathrm{base}}^{\mathrm{tgt}} \rightarrow
 L_{\mathrm{plant}}^{\mathrm{tgt}}$: \\
fre $\rightarrow$ \{ger, ned, rus, zho\} \\
ita $\rightarrow$ \{ger, ned, rus, zho\} \\
vie $\rightarrow$ \{ger, ned, rus, zho\} \\
\midrule
\textbf{SVO} \hfill \textbf{36 triplets} \\
\midrule
$L_{\mathrm{src}} \in \{$eng, jap, rus, tur, zho$\}$ \\
$L_{\mathrm{base}}^{\mathrm{tgt}} \rightarrow
 L_{\mathrm{plant}}^{\mathrm{tgt}}$: \\
eng $\rightarrow$ \{jap, tur\} \\
jap $\rightarrow$ \{eng, rus, zho\} \\
rus $\rightarrow$ \{jap, tur\} \\
tur $\rightarrow$ \{eng, rus, zho\} \\
zho $\rightarrow$ \{jap, tur\} \\
\midrule
\textbf{MV} \hfill \textbf{12 triplets} \\
\midrule
$L_{\mathrm{src}} \in \{$eng, ger, tur, zho$\}$ \\
$L_{\mathrm{base}}^{\mathrm{tgt}} \rightarrow
 L_{\mathrm{plant}}^{\mathrm{tgt}}$: \\
eng $\rightarrow$ \{tur\} \\
ger $\rightarrow$ \{tur\} \\
tur $\rightarrow$ \{eng, ger, zho\} \\
zho $\rightarrow$ \{tur\} \\
\bottomrule
\end{tabularx}
\caption{Language configurations used in the experiments.}
\label{tab:language-triplets}
\end{table}

\section{Other}

A very noticeable trait LLMs presented is their sensitivity to how the prompt is phrased. Although we commit to one fixed format, it is important to remember that the LLMs' performance is very variable. For example, we noticed that the inclusion of the phrase in the instruction significantly raises the prediction accuracy.

Below are descriptions of additional experiments, which serve as either preliminary measures, sanity checks, or out-of-scope ideas that are worth considering for future work.

\subsection{Entropy}

\begin{figure}
    \centering
    \includegraphics[width=0.75\linewidth]{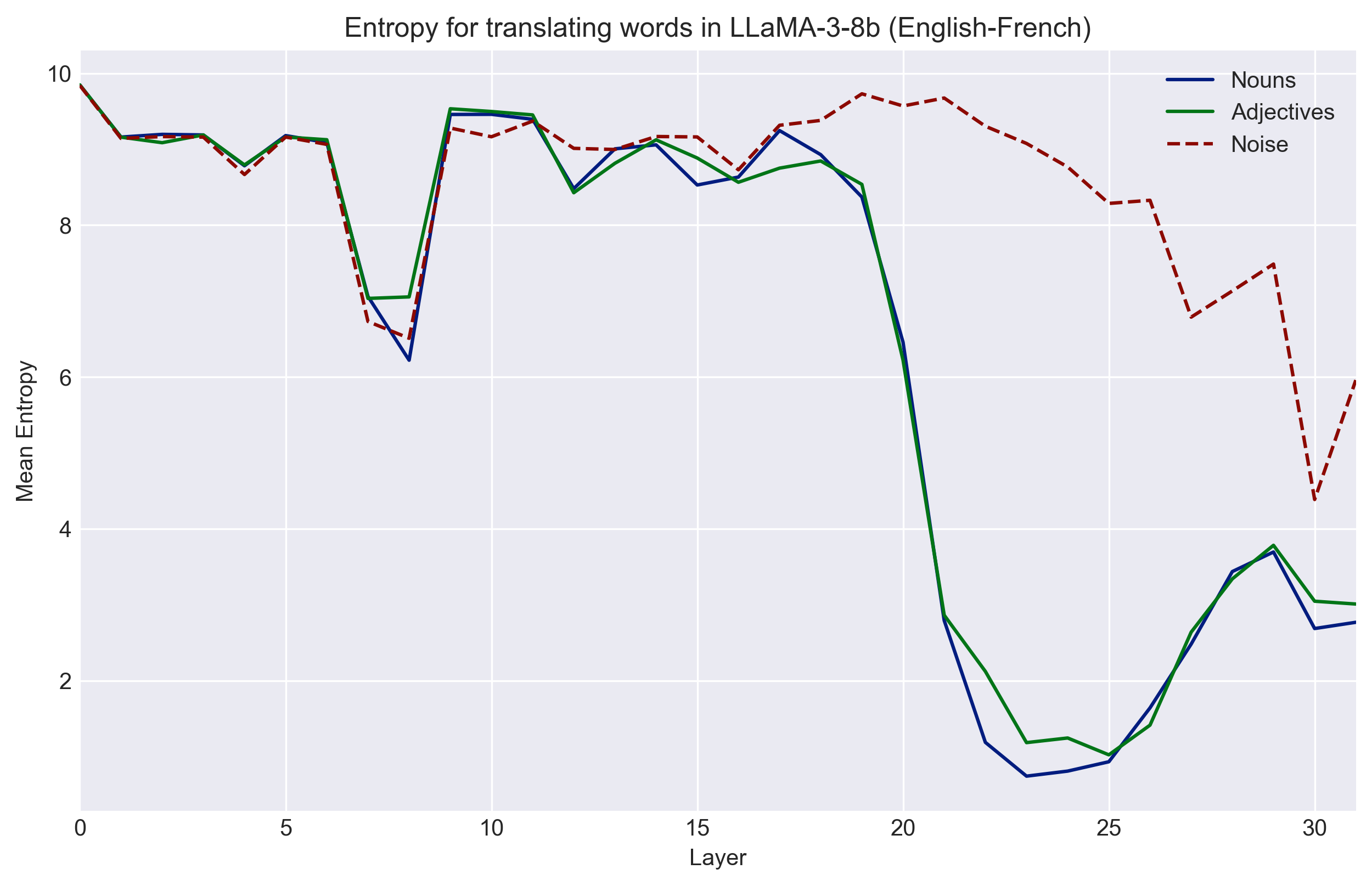}
    \caption{Mean Entropy per layer for translating an English word into French.}
    \label{fig:mean-entropy}
\end{figure}

Something worth considering in the framework of our experiments is whether different parts of speech are treated differently by the model in a translation task.
Looking at entropy at each layer, it seems that nouns and adjectives are treated rather similarly, i.e. at every stage Llama 3 is equally as sure/unsure of the upcoming token, regardless of part of speech. Comparing to noisy data (random sequence of characters of length between 3 and 8), we can deduce that there is also a difference between legitimate words and noisy input (Figure \ref{fig:mean-entropy}).

Prompt template for this calculation reads as follows:

\texttt{INSTRUCTION: Translate this word from English to French. \\
English: "\textcolor{blue}{\{word\}}". \\
French: "}

\subsection{Patching and LogitLens}

\begin{figure*}
    \begin{subfigure}[t]{0.5\textwidth}
    \centering
    \includegraphics[width=\linewidth]{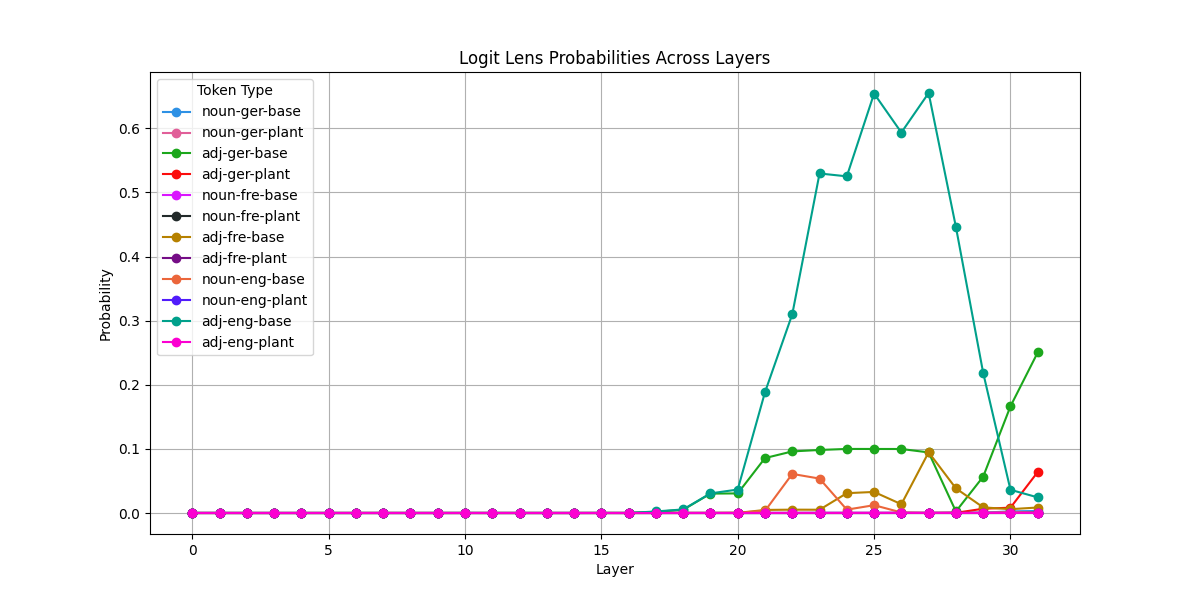}
    \caption{Patched at Layer 0}
    % \label{fig:patched_logit_lens_0}
    \end{subfigure}
    \begin{subfigure}[t]{0.5\textwidth}
    \centering
    \includegraphics[width=\linewidth]{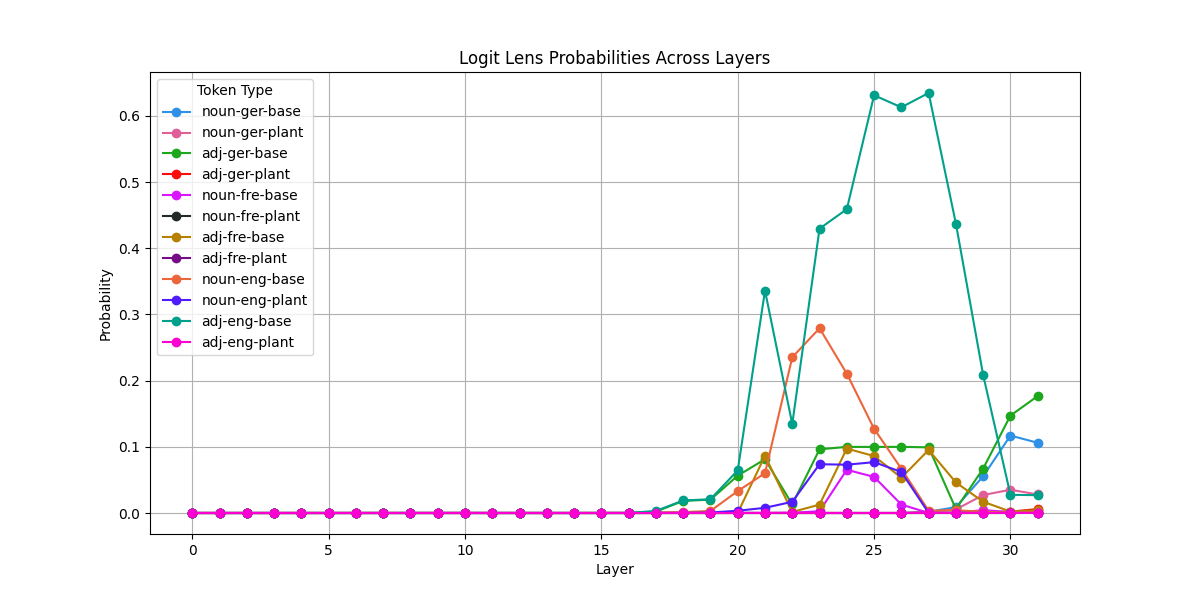}
    \caption{Patched at Layer 14}
    \end{subfigure}
     \begin{subfigure}[t]{0.5\textwidth}
    \centering
    \includegraphics[width=\linewidth]{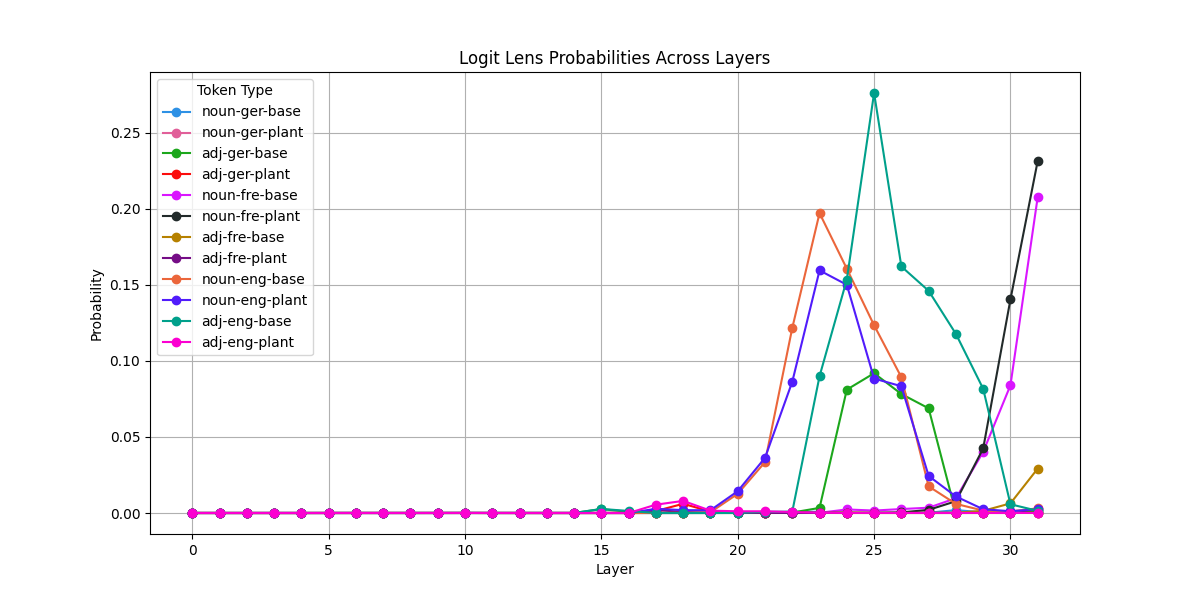}
    \caption{Patched at Layer 17}
    \end{subfigure}
    \begin{subfigure}[t]{0.5\textwidth}
    \centering
    \includegraphics[width=\linewidth]{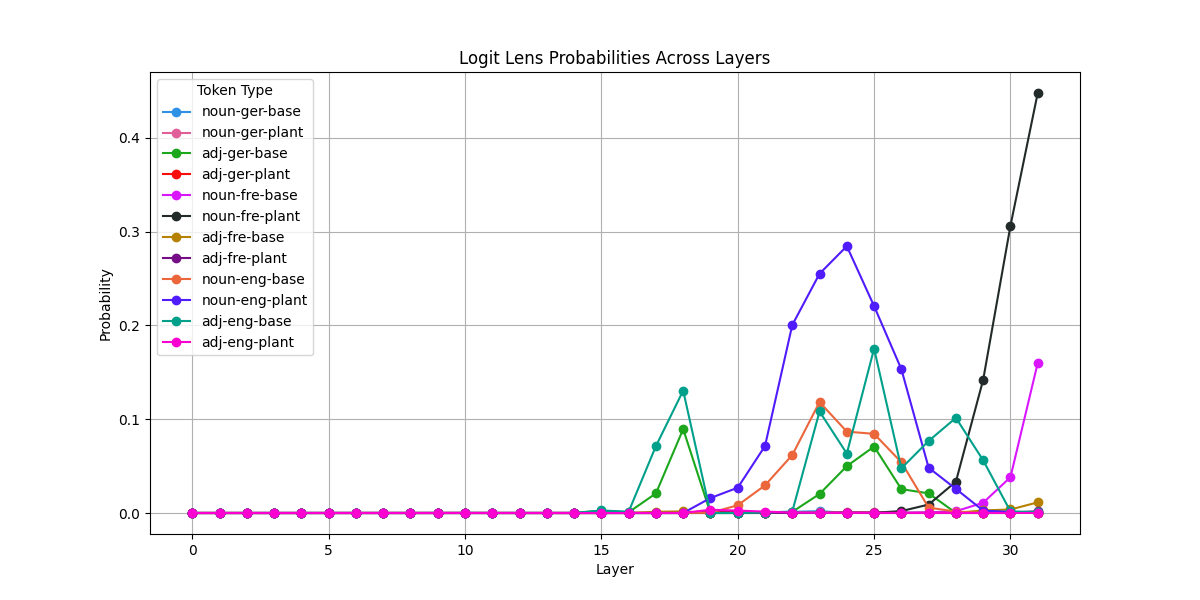}
    \caption{Patched at Layer 19}
    \end{subfigure}
    \caption{LogitLens projections of patched Llama 3 for the NP dataset. $L^\text{src} = \text{English}$, $L^\text{tgt}_\text{base} = \text{German}$, $L^\text{tgt}_\text{plant} = \text{French}$}
    \label{fig:patched_logitlens}
\end{figure*}

Fusing the experiments described in \ref{section:LogitLens} and \ref{sec:layerpatch}, we use LogitLens to look at how patching certain layers in Llama 3 affects its inner representation (Figure \ref{fig:patched_logitlens}) for one language setup that we observed: $L^\text{src} = \text{English}$, $L^\text{tgt}_\text{base} = \text{German}$, $L^\text{tgt}_\text{plant} = \text{French}$.

Crucially, we see that when patching Layer 14, there is a noticeable spike of the English noun in the inner layers, which corresponds to the plant's (French) syntax. Furthermore, the expected German adjective starts competing with the German noun. This can be regarded as evidence that Layer 14 holds information on syntax that is unrelated to a specific surface language. Something else worth noting is that patching Layer 17 leads to a higher probability of the French plant lexical component, even though our patching results suggest that Llama resolves that aspect later in the architecture.

\section{Supplementary results: LogitLens}
\label{appendix:logitlens}

Below are the plots of projected probabilities that were calculated for various configurations of (a) the language pair and (b) the dataset.

\begin{figure*}[h]
    \centering
    \resizebox{\textwidth}{!}{
    \includegraphics[width=0.8\textheight]{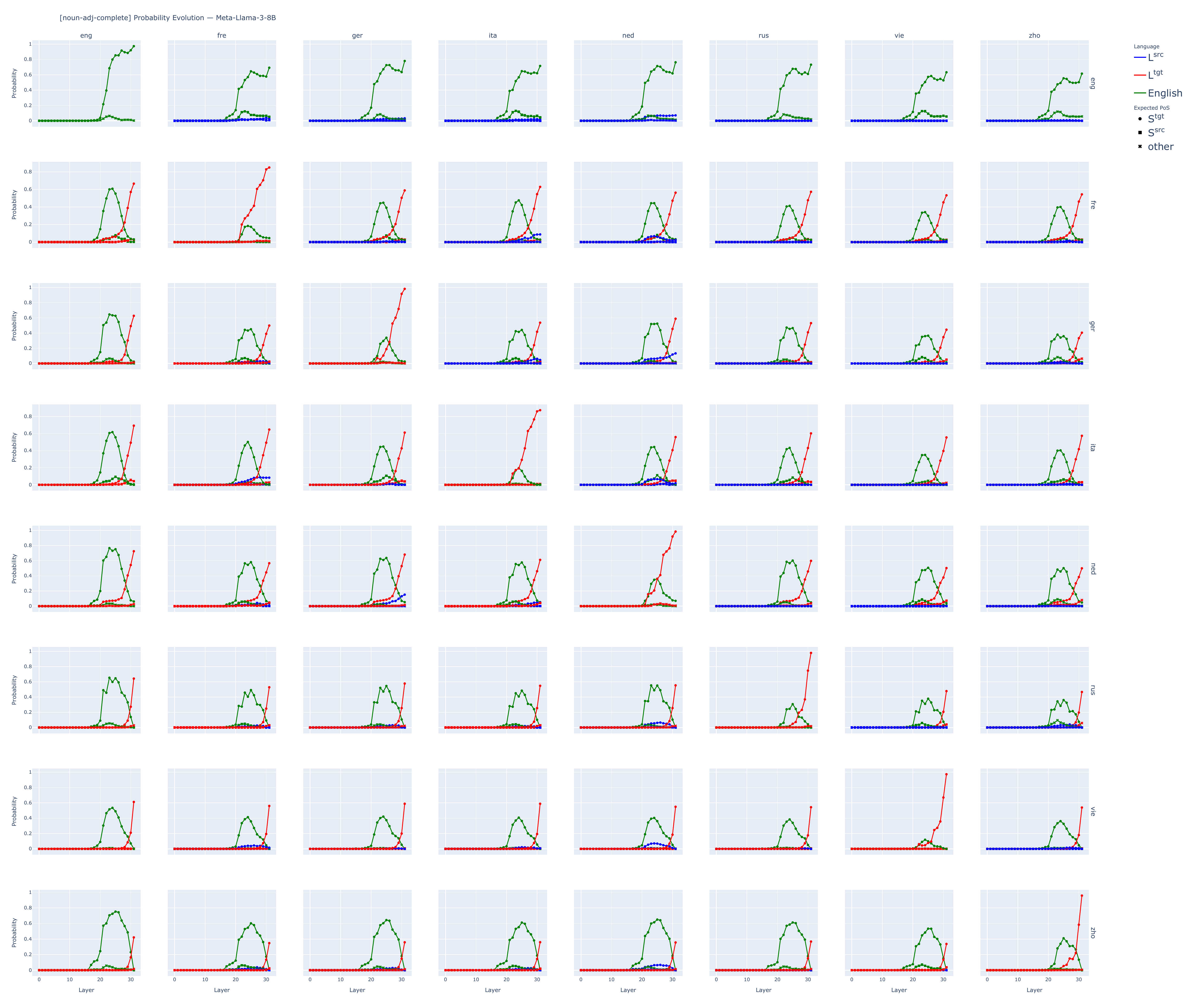}}
    \caption{Llama 3's probability of tokens throughout layers as projected by LogitLens for the NP dataset.}
    \label{fig:logitlens_noun-adj-complete_Meta-Llama-3_grid}
\end{figure*}

\begin{figure*}[h]
    \centering
    \resizebox{\textwidth}{!}{
    \includegraphics[width=0.8\textheight]{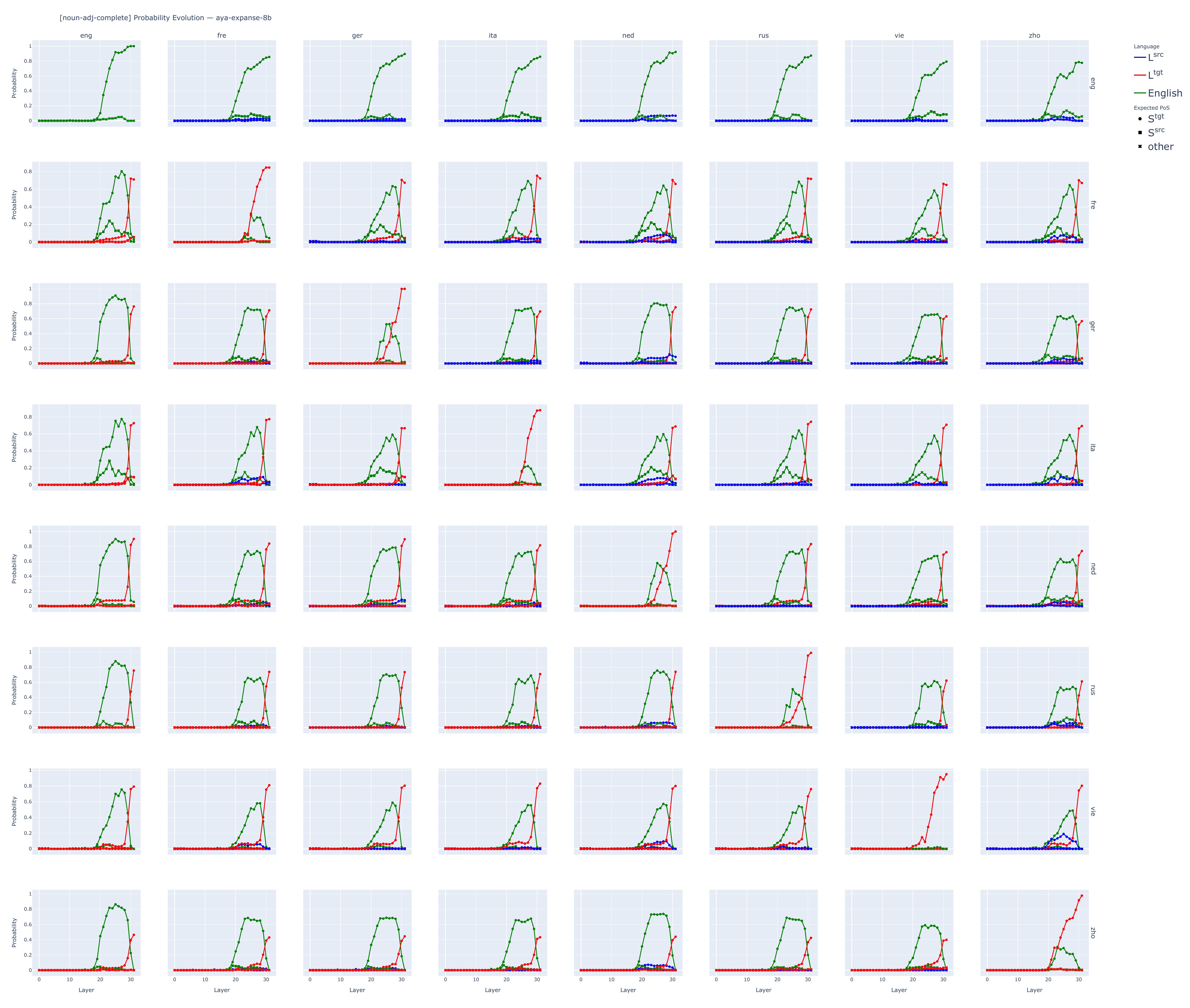}}
    \caption{Aya Expanse's probability of tokens throughout layers as projected by LogitLens for the NP dataset.}
    \label{fig:logitlens_noun-adj-complete_aya-expanse-8b_grid}
\end{figure*}

\begin{figure*}[h]
    \centering
    \resizebox{\textwidth}{!}{
    \includegraphics[width=0.8\textheight]{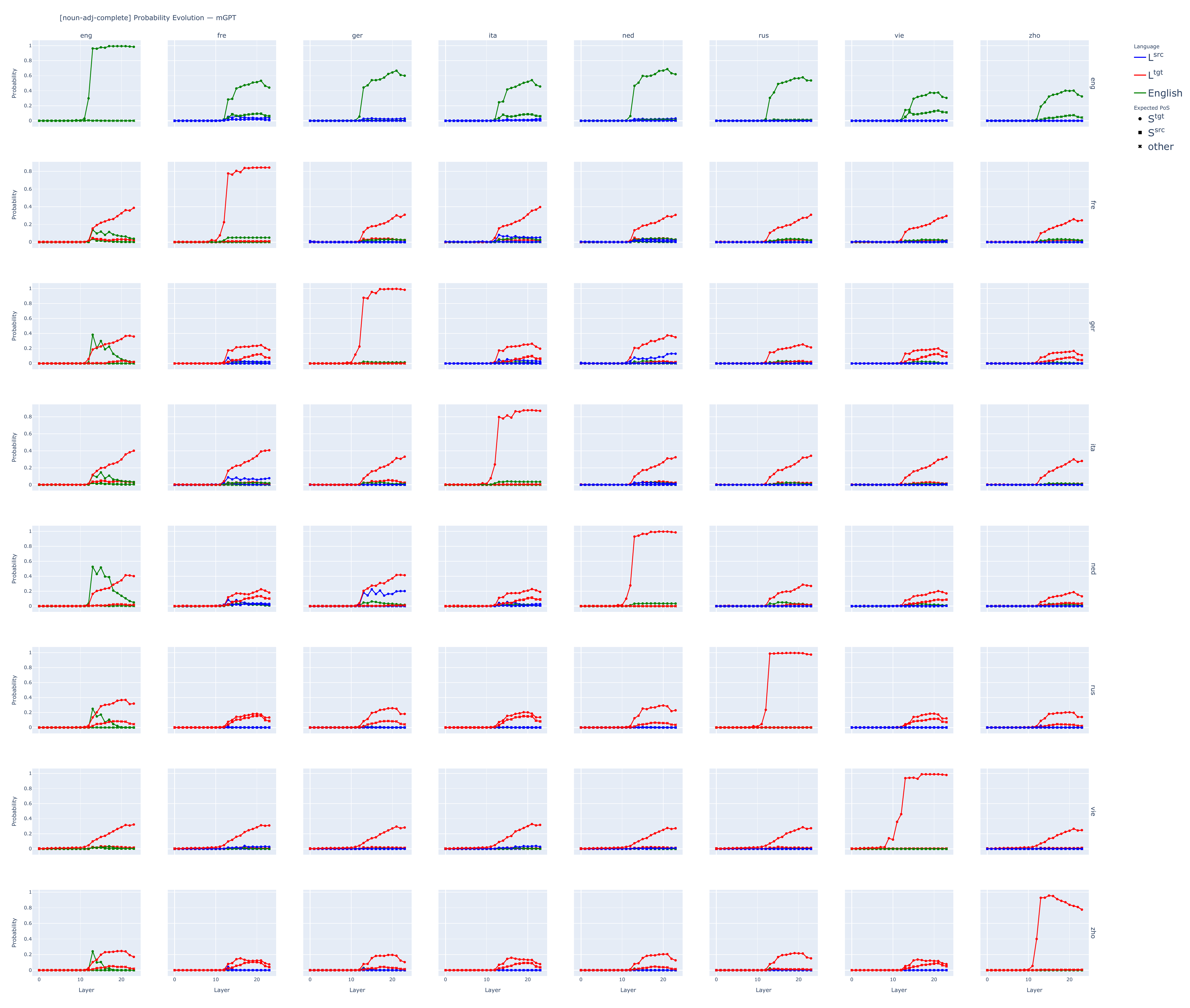}}
    \caption{mGPT's probability of tokens throughout layers as projected by LogitLens for the NP dataset.}
    \label{fig:logitlens_noun-adj-complete_mGPT_grid}
\end{figure*}

\begin{figure*}[h]
    \centering
    \resizebox{\textwidth}{!}{
    \includegraphics[width=0.8\textheight]{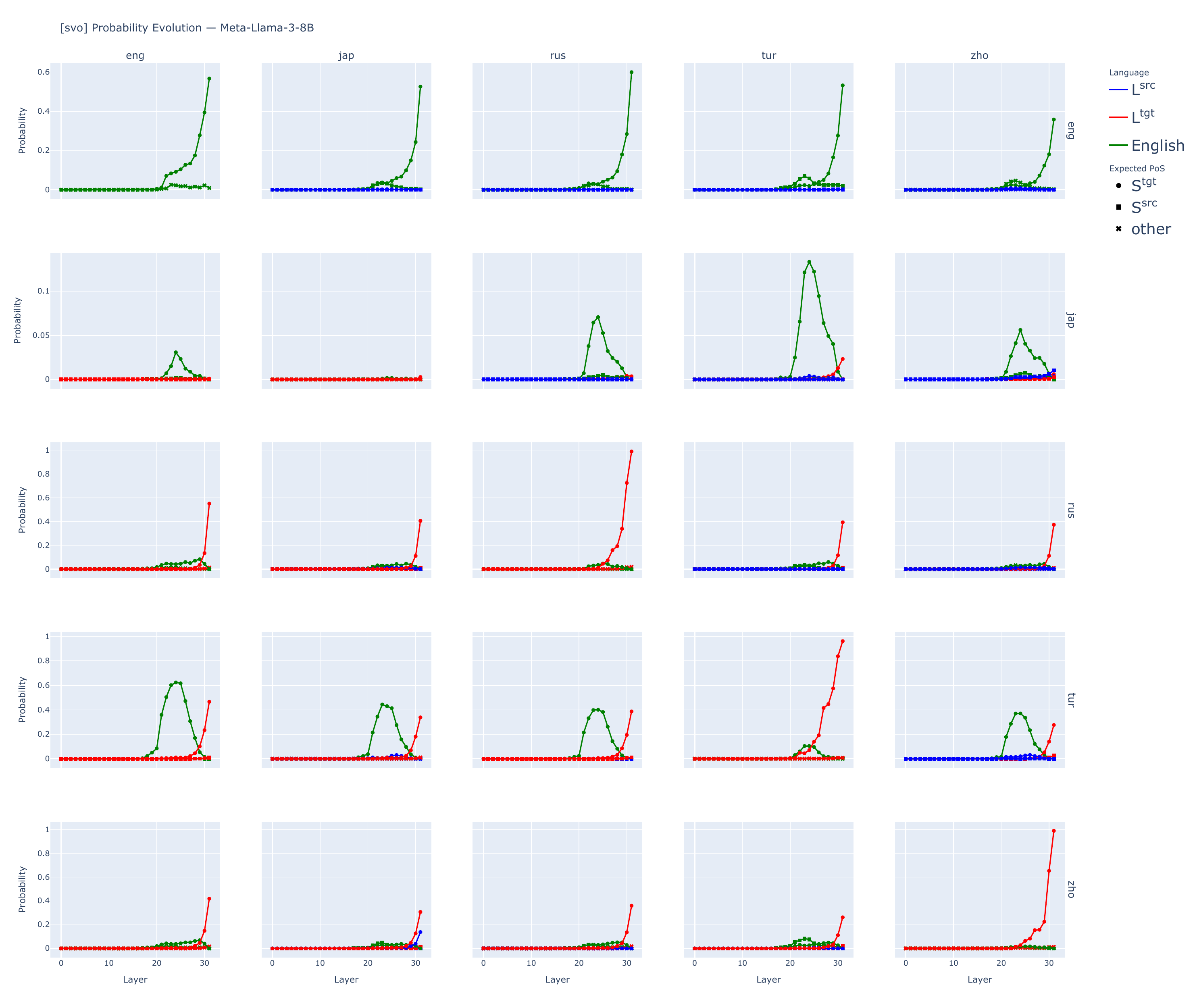}}
    \caption{Llama's probability of tokens throughout layers as projected by LogitLens for the SVO dataset.}
    \label{fig:logitlens_svo_Meta-Llama-3-8B_grid}
\end{figure*}

\begin{figure*}[h]
    \centering
    \resizebox{\textwidth}{!}{
    \includegraphics[width=0.8\textheight]{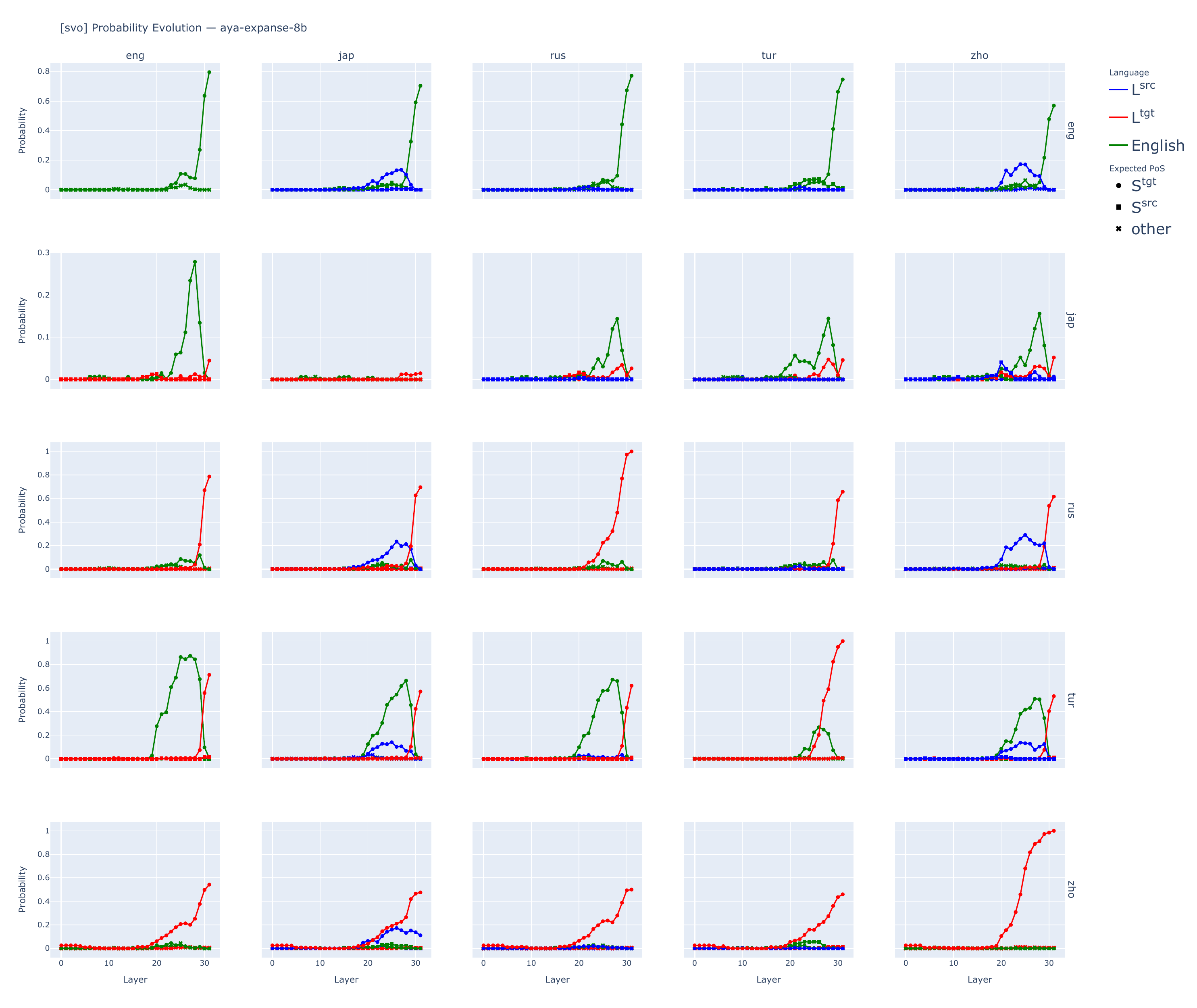}}
    \caption{Aya-Expanse's probability of tokens throughout layers as projected by LogitLens for the SVO dataset.}
    \label{fig:logitlens_svo_aya-expanse-8b_grid}
\end{figure*}

\begin{figure*}[h]
    \centering
    \resizebox{\textwidth}{!}{
    \includegraphics[width=0.8\textheight]{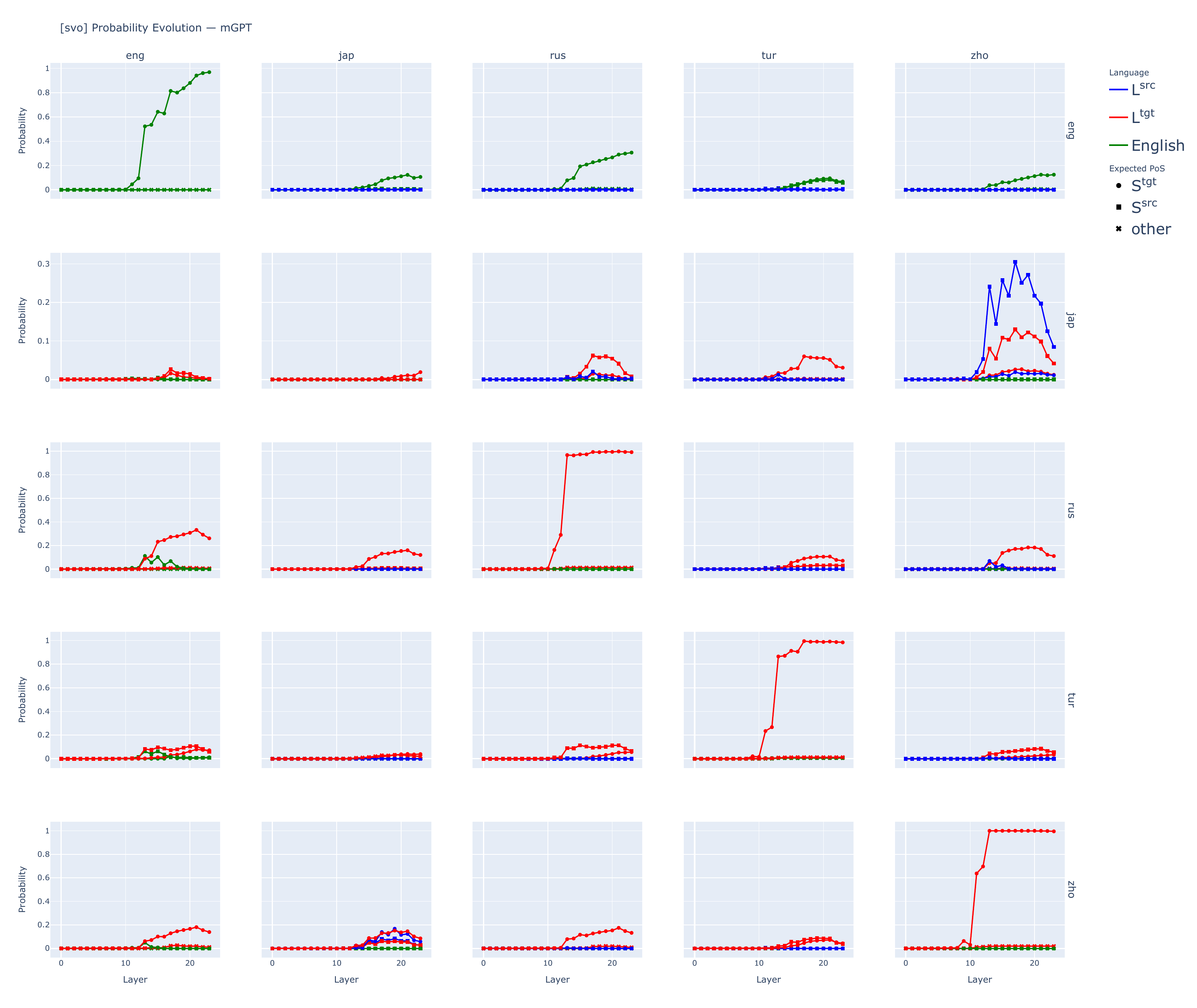}}
    \caption{mGPT's probability of tokens throughout layers as projected by LogitLens for the SVO dataset.}
    \label{fig:logitlens_svo_mGPT_grid}
\end{figure*}
\section{Supplementary Results: Mean activation Patching of Individual Attention Heads}
\label{appendix:supp_mean_head_patch}

\begin{figure}
    \centering
    \includegraphics[width=\linewidth]{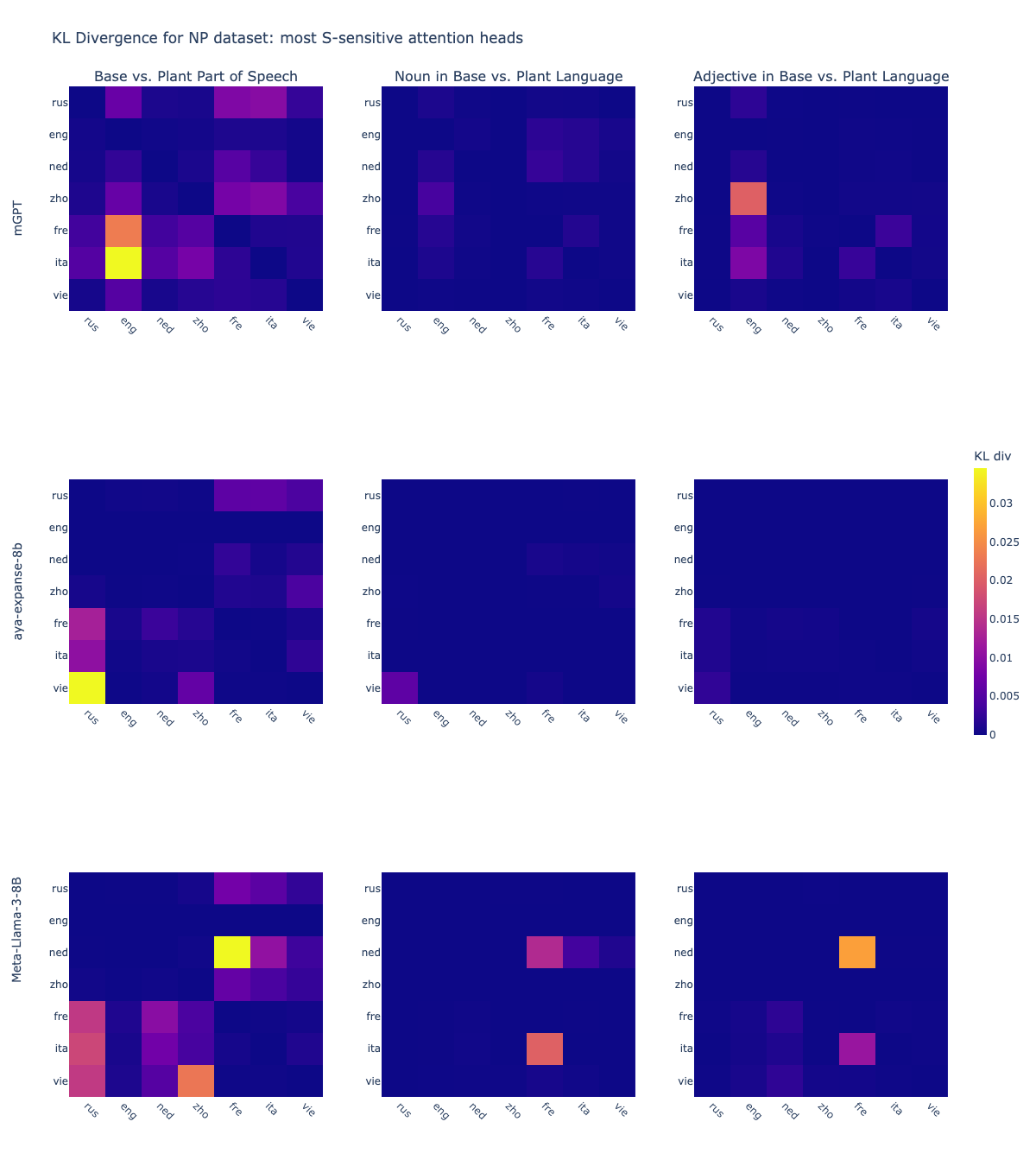}
    \caption{KL Divergences for each setting of the steering experiment with German as the source language; NP dataset. Russian (rus), English (eng), Dutch (ned), and Chinese (zho) share the Adjective-first word order, while French (fre), Italian (ita), and Vietnamese (vie) share the Noun-first word order.}
    \label{fig:kl_noun-adj-complete_ger}
\end{figure}

With other source languages, the same pattern observed in our results is repeated: KL divergence is higher for syntactically heterogeneous language pairs (as seen for German in Figure \ref{fig:kl_noun-adj-complete_ger}). A noticeable difference, however, is that English seems to follow the same pattern as German, only both ways: whether as a base or patch target language, English does not seem to influence the probabilities proportionally for Aya-Expanse or Llama.
This is easier to make sense of: English plays a seemingly different role compared to other languages.
In combination with results of our LogitLens analysis, one can theorize that English has a status of the language that represents meaning in a more neutral way, which is why it is represented by inner layer activations.
It could be that this status prevents English from the ability to steer an activation,  as well as be steered by one, too drastically.

\begin{figure}
    \centering
    \includegraphics[width=\linewidth]{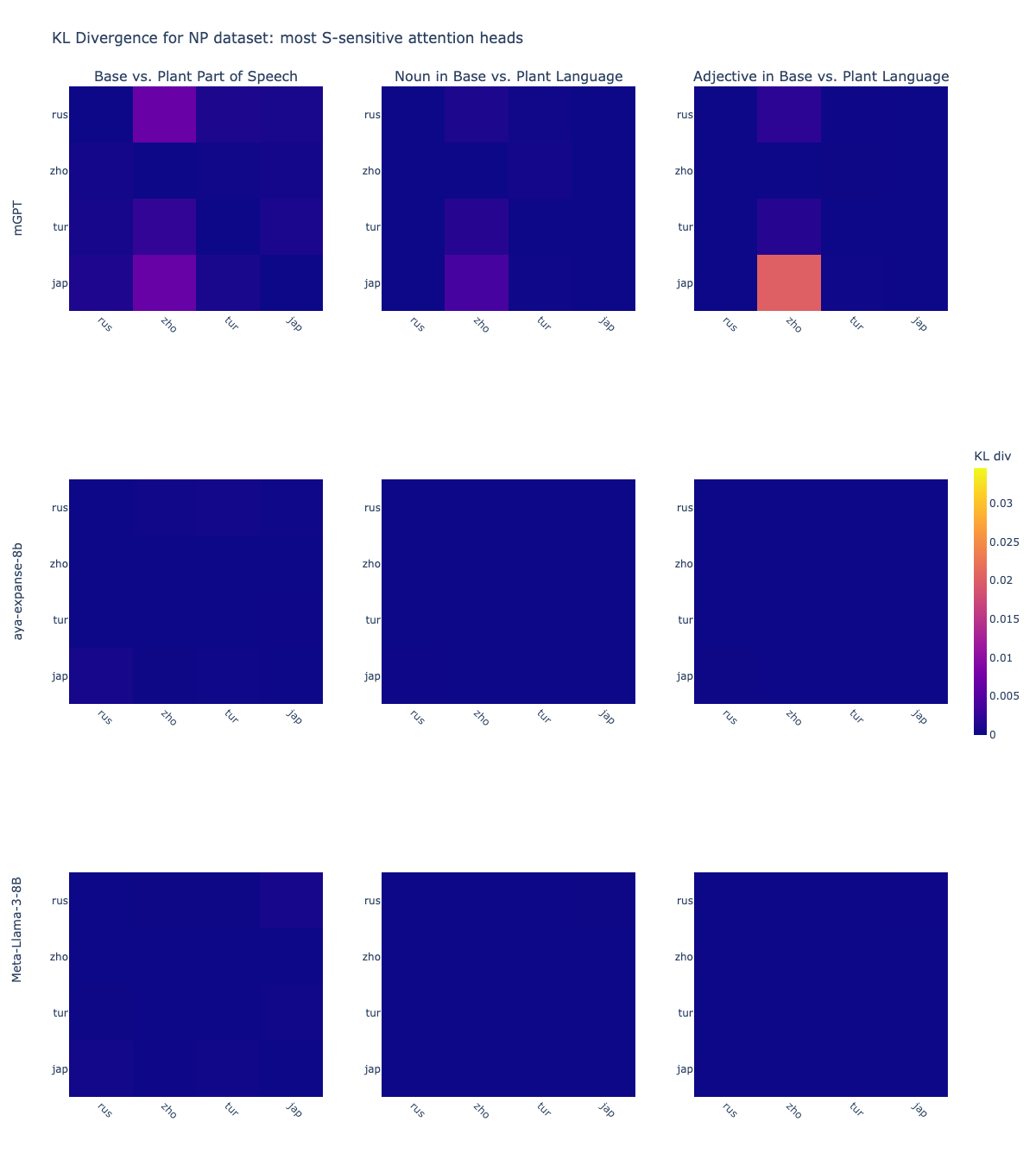}
    \caption{KL Divergences for each setting of the steering experiment with English as the source language. SVO dataset. Russian (rus) and Chinese (zho) share the SVO word order, while Turkish (tur) and Japanese (jap) share the SOV word order.}
    \label{fig:kl_svo_eng}
\end{figure}

\section{Supplementary Results: Activation Patching of Individual layers}
\label{appendix:supp_layer_patching}

\begin{table*}[th]
\centering
{\small
\begin{tabular}{ll|cc|cc|cc|c}
\toprule
& & \multicolumn{2}{c}{$(\textbf{S}_\text{plant}, L_\text{base}, C_\text{base})$} & \multicolumn{2}{c}{$(\textbf{S}_\text{plant}, \textbf{L}_\text{plant}, C_\text{base})$} & \multicolumn{2}{c}{$(\textbf{S}_\text{plant}, \textbf{L}_\text{plant}, \textbf{C}_\text{plant})$} & \multirow{2}{*}{Strategy}\\
& & $\Delta_{\max}$ & Layer & $\Delta_{\max}$ & Layer & $\Delta_{\max}$ & Layer & \\
\midrule
\multicolumn{8}{l}{\textit{SVO}} \\
\multicolumn{8}{l}{\quad $S(L_{\text{plant}})=$object, $S(L_{\text{base}})=$verb} \\
& mGPT & -0.00 & 7 & 0.02 & 12 & 0.03 & 21 & $S \rightarrow L \rightarrow C$ \\
& Aya Expanse & 0.09 & 16 & 0.15 & 19 & 0.28 & 28 & $S \rightarrow L \rightarrow C$ \\
& Llama 3 (8B) & 0.07 & 14 & 0.04 & 16 & 0.19 & 26 & $S \rightarrow L \rightarrow C$ \\
\multicolumn{8}{l}{\quad $S(L_{\text{plant}})=$verb, $S(L_{\text{base}})=$object} \\
& mGPT & -0.00 & 7 & 0.06 & 12 & 0.15 & 19 & $S \rightarrow L \rightarrow C$ \\
& Aya Expanse & 0.10 & 13 & 0.43 & 16 & 0.53 & 27 & $S \rightarrow L \rightarrow C$ \\
& Llama 3 (8B) & 0.09 & 14 & 0.19 & 14 & 0.43 & 27 & $S, L \rightarrow C$ \\
\midrule
\multicolumn{8}{l}{\textit{MV}} \\
\multicolumn{8}{l}{\quad $S(L_{\text{plant}})=$modal, $S(L_{\text{base}})=$verb} \\
& mGPT & 0.01 & 7 & 0.12 & 14 & 0.61 & 23 & $S \rightarrow L \rightarrow C$ \\
& Aya Expanse & 0.00 & 14 & 0.45 & 19 & 0.68 & 29 & $S \rightarrow L \rightarrow C$ \\
& Llama 3 (8B) & 0.00 & 15 & 0.28 & 16 & 0.75 & 31 & $S \rightarrow L \rightarrow C$ \\
\multicolumn{8}{l}{\quad $S(L_{\text{plant}})=$verb, $S(L_{\text{base}})=$modal} \\
& mGPT & 0.00 & 3 & 0.03 & 12 & 0.09 & 19 & $S \rightarrow L \rightarrow C$ \\
& Aya Expanse & 0.12 & 18 & 0.36 & 20 & 0.59 & 24 & $S \rightarrow L \rightarrow C$ \\
& Llama 3 (8B) & 0.10 & 15 & 0.12 & 17 & 0.41 & 25 & $S \rightarrow L \rightarrow C$ \\
\bottomrule
\end{tabular}}
\caption{Extended activation patching results, grouped by different word order settings.}
\label{tab:extended-patching}
\end{table*}

\begin{figure}
    \centering
    \includegraphics[width=\linewidth]{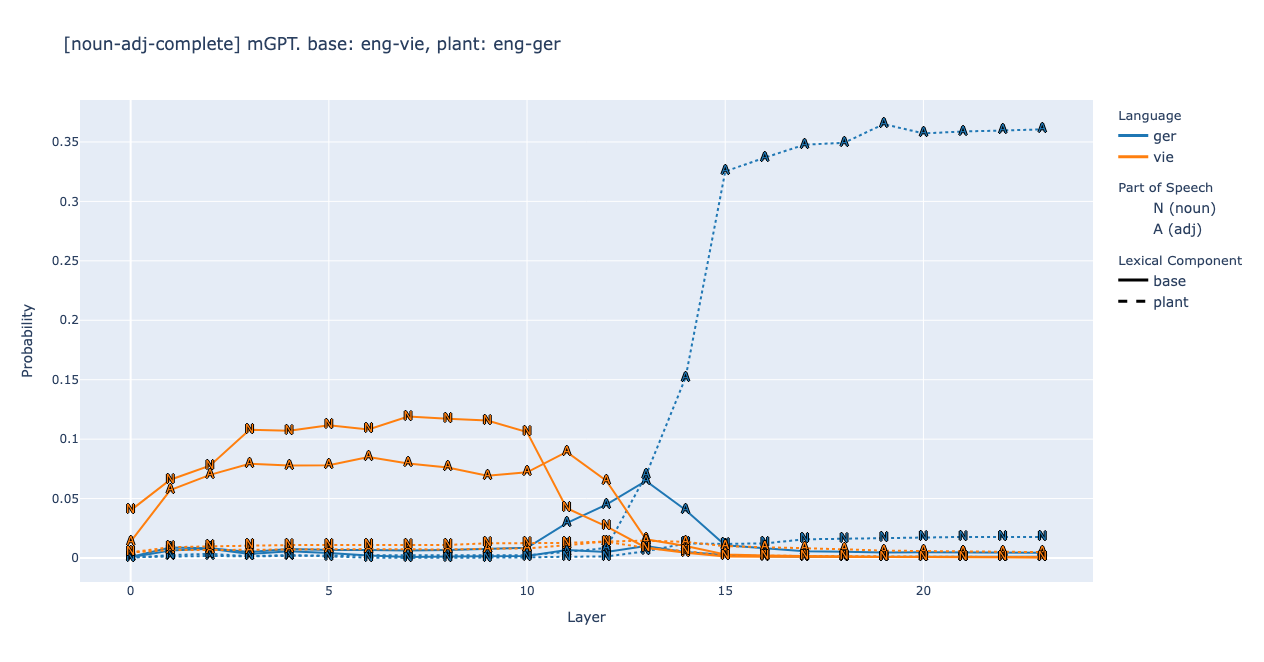}
    \caption{mGPT's probabilities of different tokens resulting from patching layer activations: $L_\text{base}$ is Vietnamese, $L_\text{plant}$ is German.}
    \label{fig:noun-adj-complete_mGPT_eng-vie_eng-ger}
\end{figure}

Another clear example of the $S \rightarrow L \rightarrow C$ strategy is shown for mGPT in Figure \ref{fig:noun-adj-complete_mGPT_eng-vie_eng-ger}, with Vietnamese (noun-first) and German (adjective-first) as the base and plant target languages, respectively.

\begin{figure}
    \centering
    \includegraphics[width=\linewidth]{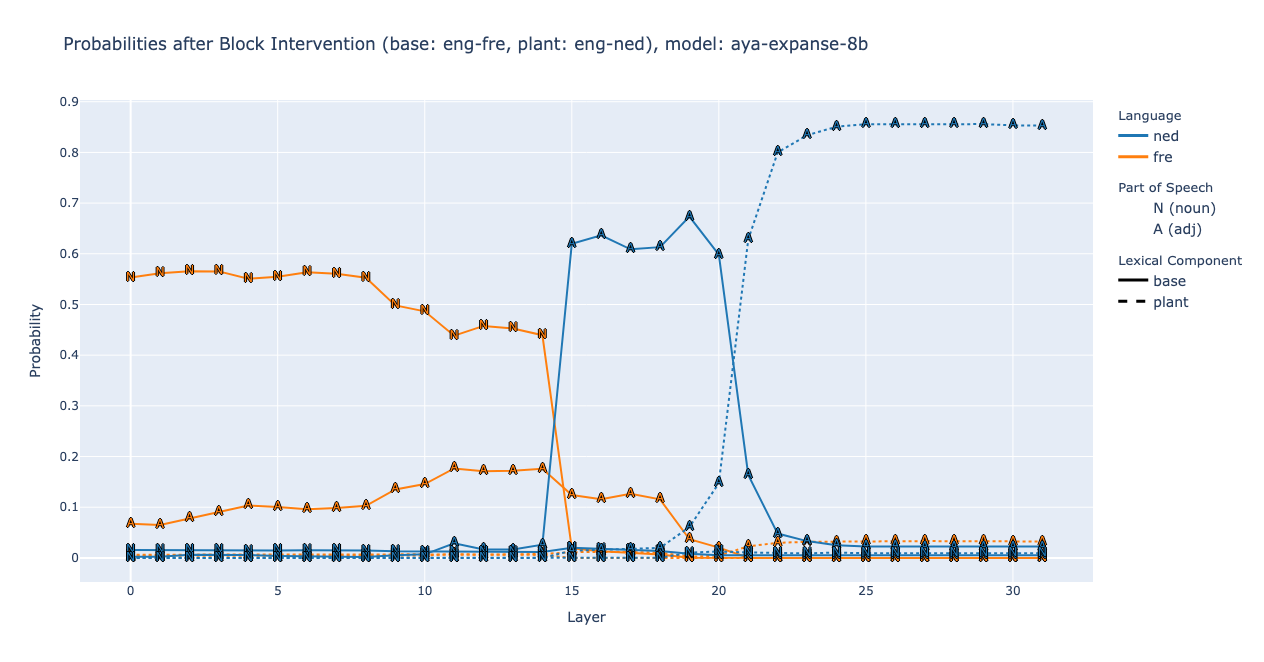}
    \caption{Aya Expanse's probabilities of different tokens resulting from patching layer activations: $L_\text{base}$ is French, $L_\text{plant}$ is Dutch.}
    \label{fig:noun-adj-complete_aya-expanse-8b_eng-fre_eng-ned}
\end{figure}

Looking at the raw probabilities in Aya Expanse, however, we notice part of speech $S$ and language $L$ ``switch'' at the same layer (Layer 15), as the token differing by only part of speech from the base token does not emerge with a high probability (Figure \ref{fig:noun-adj-complete_aya-expanse-8b_eng-fre_eng-ned}).
This suggests that the information regarding the language and syntax are tied together more closely in Aya Expanse's representations than in those of mGPT or Llama-3.
However, even in those cases we see a rise in probability of $S_\text{plant}$, just not as strong as to overcome $S_\text{base}$.
We leave further analysis for future work.

\subsection{Natural Language Data}
\label{subsection:natual-language-data}

We gathered 27 sentences from the multilingual parallel dataset FLORES-200 \cite{nllbteam2022languageleftbehindscaling} and annotated the English, German, and French translations for noun phrases with adjectives, which provides us with 8 possible language setups.

The aggregated results of our patching experiments for the dubbed NP Natural dataset are shown in Table \ref{tab:delta-table_np-natural}.
We see potential evidence for the $S \rightarrow L \rightarrow C$ strategy for mGPT and Aya Expanse, whereas Llama 3 does not exhibit it at first glance.
However, given the size of the dataset, it is difficult to draw any substantial conclusions.
An examination of whether more naturalistic language data is treated differently in terms of syntax and surface language is worth pursuing for future interpretability research.

\begin{table*}[t]
    \centering
    \begin{tabular}{ll|cc|cc|cc|cc|c}
    \toprule
     & & \multicolumn{2}{c}{$(\textbf{S}_\text{plant}, L_\text{base}, C_\text{base})$} & \multicolumn{2}{c}{$(\textbf{S}_\text{plant}, \textbf{L}_\text{plant}, C_\text{base})$} & \multicolumn{2}{c}{$(\textbf{S}_\text{plant}, \textbf{L}_\text{plant}, \textbf{C}_\text{plant})$} & \multirow{2}{*}{Strategy}\\
     & & $\Delta_{\max}$ & Layer & $\Delta_{\max}$ & Layer & $\Delta_{\max}$ & Layer &  \\
     \midrule
     & mGPT & -0.04 & 3 & 0.02 & 11 & 0.41 & 21 & $S \rightarrow L \rightarrow C$\\
     & Aya Expanse & 0.09 & 11 & 0.05 & 16 & 0.66 & 30 & $S \rightarrow L \rightarrow C$\\
     & Llama 3 (8B) & 0.04 & 14 & 0.01 & 14 & 0.56 & 31 & $S, L \rightarrow C$\\
    \bottomrule
    \end{tabular}
    \caption{Highest probability difference between the patched and clean runs for different language aspect combinations and the corresponding patched layers, averaged across all languages for the \textit{NP Natural} dataset.}
    \label{tab:delta-table_np-natural}
\end{table*}

\end{document}